\documentclass[10pt,twocolumn,letterpaper]{article}

\PassOptionsToPackage{dvipsnames,table}{xcolor}

\usepackage{cvpr}              
\usepackage{amsmath}
\usepackage{amssymb}
\usepackage{multirow}
\usepackage{algorithmic}

\usepackage{booktabs}
\usepackage{verbatim}
\usepackage{pifont}
\usepackage{ifsym}
\usepackage{graphicx}

\usepackage[euler]{textgreek}
\usepackage{arydshln}
\usepackage{marvosym}
\usepackage[accsupp]{axessibility}

\usepackage{stfloats}
\usepackage{cuted}

\newcommand{\Ours}{DeCo-Diff{}}
\newcommand{\MVTEC}{MVTec-AD{}}

\newcommand{\xx}{\boldsymbol{x}}
\newcommand{\zz}{\boldsymbol{z}}
\newcommand{\aaa}{\boldsymbol{a}}
\newcommand{\ee}{\boldsymbol{\epsilon}}
\newcommand{\etabold}{\boldsymbol{\eta}}

\newcommand{\mypar}[1]{\noindent\textbf{#1}~}
\newcommand{\colour}[2]{\textcolor{#1}{\textbf{#2}}}

\newcommand{\phz}{\phantom{0}}

\newcommand{\pixperf}[4]{{\colorbox{gray!15}{#1}\,/\,#2\,/\,\colorbox{gray!15}{#3}\,/\,#4}}
\newcommand{\imgperf}[3]{{\colorbox{gray!15}{#1}\,/\,#2\,/\,\colorbox{gray!15}{#3}}}
\newcommand{\xmark}{\ding{55}}%

\newcommand{\first}[1]{\colour{blue}{#1}}
\newcommand{\second}[1]{\colour{red}{#1}}

\definecolor{cvprblue}{rgb}{0.21,0.49,0.74}
\usepackage[pagebackref,breaklinks,colorlinks,allcolors=cvprblue]
{hyperref}

\usepackage{xspace}

\usepackage{color} 
\definecolor{darkblue}{rgb}{0.21,0.49,0.74}
\newcommand{\appensecref}[1]{\textcolor{darkblue}{Appendix~\ref{#1}}\xspace}

\definecolor{Gray}{rgb}{0.5,0.5,0.5}
\definecolor{Better}{rgb}{0.18, 0.407, 0.266}
\definecolor{Worse}{rgb}{0.35, 0.35, 0.35}

\newcommand{\imp}[1]{$_{{\textbf{\textcolor{Gray}{#1}}}}$}

\definecolor{mygreen}{RGB}{38, 199, 149}

\def\paperID{10436} 
\def\confName{CVPR}
\def\confYear{2025}

\title{Correcting Deviations from Normality: A Reformulated Diffusion Model for Multi-Class Unsupervised Anomaly Detection}

\author{Farzad Beizaee$^{1,2}$\thanks{Corresponding author: \href{mailto:farzad.beizaee.1@ens.etsmtl.ca}{farzad.beizaee.1@ens.etsmtl.ca}} \qquad Gregory A. Lodygensky$^{2}$ \qquad Christian Desrosiers$^{1}$ \qquad Jose Dolz$^{1}$ \\
$^1$ÉTS Montreal \qquad
$^2$CHU-Sainte-Justine Montreal}

\begin{document}
\maketitle
\begin{abstract}
Recent advances in diffusion models have spurred research into their application for Reconstruction-based unsupervised anomaly detection. However, these methods may struggle with maintaining structural integrity and recovering the anomaly-free content of abnormal regions, especially in multi-class scenarios. Furthermore, diffusion models are inherently designed to generate images from pure noise and struggle to selectively alter anomalous regions of an image while preserving normal ones. This leads to potential degradation of normal regions during reconstruction, hampering the effectiveness of anomaly detection. 
This paper introduces a reformulation of the standard diffusion model geared toward selective region alteration, allowing the accurate identification of anomalies.
By modeling anomalies as noise in the latent space, our proposed \textbf{Deviation correction diffusion} (\Ours) model preserves the normal regions and encourages  
transformations exclusively on anomalous areas. This selective approach enhances the reconstruction quality, facilitating effective unsupervised detection and localization of anomaly regions. Comprehensive evaluations demonstrate the superiority of our method in accurately identifying and localizing anomalies in complex images, with pixel-level AUPRC improvements of 11-14\% over state-of-the-art models on well-known anomaly detection datasets.
The code is available at \href{https://github.com/farzad-bz/DeCo-Diff}{https://github.com/farzad-bz/DeCo-Diff}
\end{abstract}    
\begin{figure}[!ht]
\centering
\includegraphics[width=\linewidth]{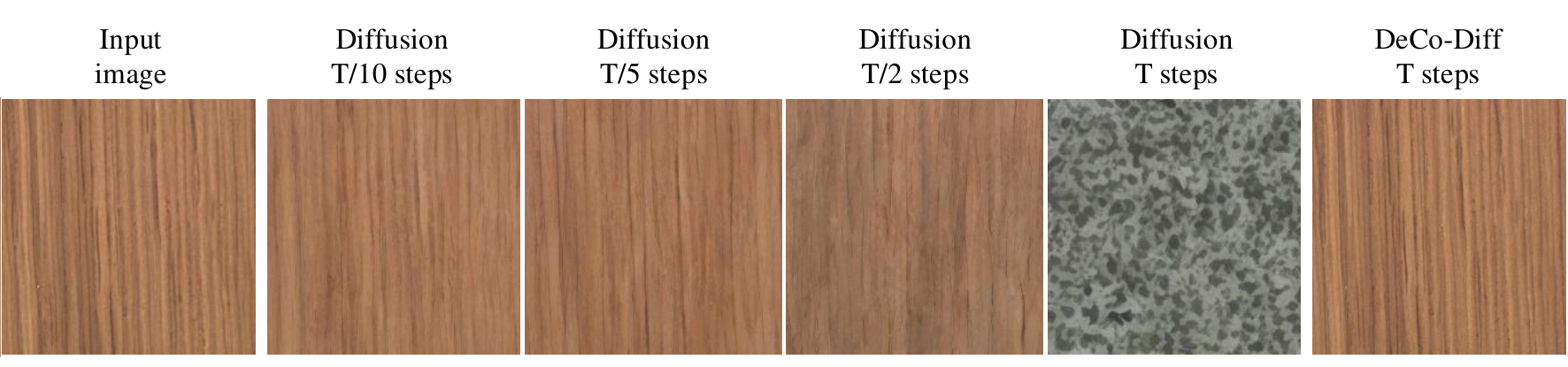}
     \caption{\textbf{Diffusion model reconstruction vs. \Ours{}.} Fine details and patterns of a normal image are 
     changed during the standard forward-backward diffusion process: the ``wood" image becomes a ``Tile" sample when $T$ steps are applied. In contrast, \Ours{} does not alter the details of the image using $T$ correction steps, maintaining the appearance of the original input image.}
\label{fig:diff_vs_deco}
\end{figure}

\begin{figure*}[!ht]
\centering
\includegraphics[width=\linewidth]{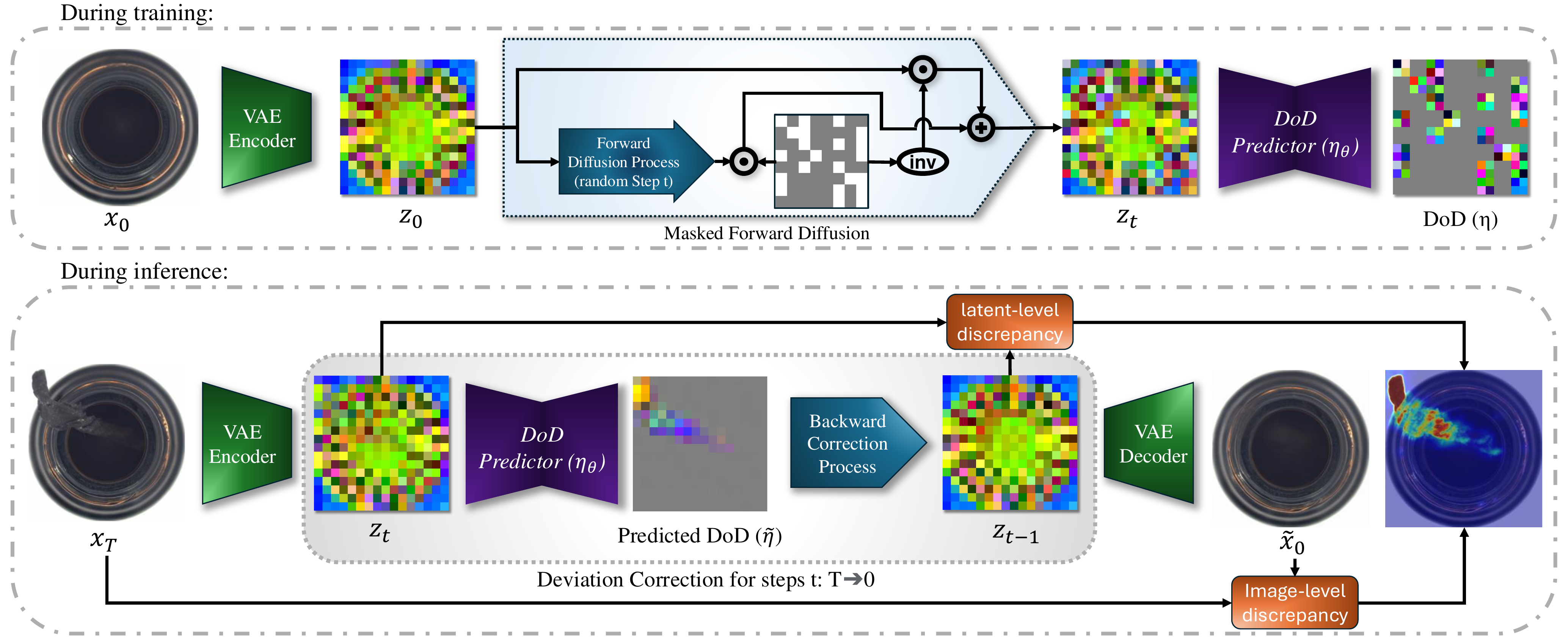}
     \caption{\textbf{Overview of the proposed method.} During \textbf{training} (\textit{top}), normal images are partially diffused using random masks and randomly sampled time-steps ([1,$T$]). Then, our \Ours{} model is trained to predict the direction of deviation from the input image. At \textbf{inference} (\textit{bottom}) starting from time-step $T$ for the target images, \Ours{} progressively corrects the deviation from normality.}
\label{fig:method}
\end{figure*}

\section{Introduction}

Unsupervised anomaly detection (UAD) is paramount to a wide span of computer vision problems across strategic and high-impact domains, such as industrial inspection \cite{defard2021padim,liu2023simplenet}, video surveillance \cite{pang2020self,yan2023feature}, or medical imaging \cite{schlegl2017unsupervised,silva2022constrained}. The main goal of UAD is to identify corrupted images, as well as anomalous pixels within these images, by leveraging only \textit{normal} images. This setting naturally arises in many scenarios, where normal samples are readily available but compiling a curated set of labeled abnormal images is costly, due to the complexity of the annotation process as well as the scarcity and high variability of potential abnormalities. 

The prevailing approach to address UAD frames this task as a \textit{cold-start}\footnote{Also commonly known as one-class classification (OCC).} anomaly detection problem, which is typically approached through three primary categories of methods: reconstruction \cite{yan2021learning,perera2019ocgan,schlegl2017unsupervised}, synthesizing \cite{li2021cutpaste,zavrtanik2021draem,schluter2022natural,zavrtanik2022dsr} and embedding \cite{defard2021padim,deng2022anomaly,roth2022towards,rudolph2022fully} based approaches. While popular, single-class approaches 
hinder the scalability of these strategies, as the amount of storage and training time increases with the number of categories. Thus, there is a real need for novel methods that can accommodate the multi-class scenario in a robust and efficient manner. 

Under the unified scenario, however, the distribution of normal data becomes more complex. As a result, the success of this task hinges on robust models capable of effectively learning the joint distribution across diverse types of objects. Diffusion models \cite{sohl2015deep}, and more particularly Denoising Diffusion Probabilistic Models (DDPM) \cite{ho2020denoising}, have emerged as strong candidates for this endeavor. Indeed, diffusion models have demonstrated strong potential in reconstructing the normal counterpart of an image to localize anomalous regions \cite{fuvcka2025transfusion,zhang2023unsupervised}. 

Diffusion models, though highly effective at generating high-quality samples, are designed to generate images from pure noise. Therefore, their direct application to modify the input images necessitates a forward diffusion process followed by a backward denoising process. However, this strategy imposes several limitations. First of all, directly using DDPMs in the multi-class context may result in misclassifying generated images due to the loss of their original category information \cite{he2024diffusion}, when a high number of diffusion steps is applied. On the other hand, if we use lower diffusion steps, the \textit{``identity shortcut''} issue will emerge, where the model tends to simply denoise the input regardless of whether the content is normal or anomalous, thus preserving the anomalous regions\cite{you2022unified}. A further problem arises when forward and reverse diffusion steps are applied indiscriminately across the entire image. In such cases, the model may struggle to fully recover the normal patterns leading to false identification of anomalies due to discrepancies between the input image and its reconstruction. This problem is more significant when the input image involves random patterns and textures, such as ``tile". Lastly, the model's ability to accurately reconstruct the normal appearance of anomalous regions improves by leveraging the contextual information from surrounding normal regions. However, diffusion models introduce noise across the entire image, leading to the loss of crucial details required for reconstructing the normal counterpart of the image. These limitations are depicted in \cref{fig:diff_vs_deco}.

To address these limitations, we propose a reformulation of the standard diffusion models in such a way that it corrects the deviation from normality. This new formulation better suits our goal of reconstructing 
abnormal regions into their normal counterpart, while preserving the fine details of normal areas. Specifically, it enables us to directly use the target anomalous image in the backward correction process during inference, which eliminates the need for forward diffusion process, prevents the degradation of informative information, and enables selective correction of abnormal regions.The proposed model also introduces a random masking strategy for the added noise, which brings two important benefits. First, it leaves portions of the image untouched during the forward process, thus making the correction process conditioned on those normal untouched regions. Secondly, during the backward correction process, using our reformulation, the model learns to selectively alter abnormal regions of the image while keeping normal areas unchanged. This leads to a more precise anomaly detection and localization.

{
Our \textbf{key contributions} can be summarized as follows:
\begin{itemize} \setlength\itemsep{.25em}
    \item We propose a novel reformulation of diffusion models that learns to correct deviation from the learned distribution of normality to its normal counterparts, rather than generating samples through denoising steps.

    \item We introduce a random masking strategy into the forward diffusion process, which conditions the deviation correction of abnormal regions to surrounding areas while preserving the fine details of normal regions.

    \item Furthermore, we reformulated the DDIM sampling \cite{songdenoising} to accommodate the deviation correction approach presented, enabling a faster, stable, and deterministic reverse process for efficient sampling.

    \item Extensive results on popular anomaly detection benchmarks demonstrate the superiority of our approach across recent state-of-the-art methods and multiple metrics. 
\end{itemize}

}

\section{Related work}

\subsection{Unsupervised Anomaly detection}

Unsupervised anomaly detection 
has been studied from multiple perspectives: \textbf{\textit{1) Reconstruction-based approaches}} rely on the assumption that a model trained on normal data will fail to reconstruct anomalous regions, leveraging the differences between the input and its reconstructed image as an anomaly score. Generative Adversarial Networks \cite{yan2021learning,perera2019ocgan,schlegl2017unsupervised}, Variational Auto-Encoders \cite{liu2020towards}, and Normalizing Flows \cite{gudovskiy2022cflow}, 
are typically used as the backbone for reconstruction networks in this task. \textbf{\textit{2) Synthesizing-based approaches}} typically introduce synthetic anomalies in normal images used for training \cite{li2021cutpaste,zavrtanik2021draem,schluter2022natural,zavrtanik2022dsr}. For example, DR\AE
M \cite{zavrtanik2021draem} trains a network on synthetically-generated just-out-of-distribution patterns, whereas CutPaste \cite{li2021cutpaste} introduces anomalies via a simple, yet efficient strategy based on cutting-pasting image patches at random locations on a normal image. \textbf{\textit{3) Embedding-based approaches}} focus on embedding the normal features into a compressed space, relying on the assumption that anomalous features are far from the normal clusters in this space. These methods employ networks that are pre-trained on large-scale datasets like ImageNet for feature extraction \cite{defard2021padim,deng2022anomaly,roth2022towards,rudolph2022fully}. For instance, PaDiM \cite{defard2021padim} resorts to a multivariate Gaussian to model the distribution of normal patch features at each position of the image, and then measures the normality score using the Mahalanobis distance. Similarly, PatchCore \cite{roth2022towards} employs a core set to encode the features of normal patches, and finds anomalies at test time by computing the distance from a new patch's embedding and its nearest element in the core set. More recently, \cite{li2024hyperbolic} proposed using the hyperbolic space to measure the distance between feature representations. Despite the progress made by these approaches in the single-class setting, their performance in the unified scenario remains unsatisfactory.

\subsection{Multi-class unsupervised anomaly detection}

The direct application of above-mentioned methods to the multi-class problem often leads to suboptimal performance. Moreover, since each category needs a separate trained model, 
the computational burden of these methods quickly explodes as the number of classes increases.

Tackling this more realistic scenario, multi-class unsupervised anomaly detection (Mc-UAD) approaches have recently gained traction within the community \cite{you2022unified,deng2022anomaly,he2024diffusion,liu2023simplenet,zhang2023destseg,he2024mambaad,zhang2023exploring,guo2024recontrast}. UniAD \cite{you2022unified} proposes a series of modifications to accommodate reconstruction-based networks for the challenging task of Mc-UAD. 
Observing the importance of the query embedding module in transformers to model the normal distribution, a query decoder is integrated into each layer, instead of only in the first layer as in vanilla transformers. To alleviate the ``\textit{identity shortcut}" problem in transformers, authors also introduce a neighbor mask attention module preventing tokens to copy themselves via self-attention. Last, a feature jittering strategy is employed to help the model recover normal features from noisy ones. Rd4AD \cite{deng2022anomaly} leverages a reverse distillation approach, where a student network learns to restore the multi-scale representations of a teacher given the teacher one-class embeddings. 
MambaAD \cite{he2024mambaad} proposed a pyramidal autencoder framework to reconstruct multi-scale features using recently proposed Vision Mamba networks \cite{zhu2024vision}. MoEAD \cite{mengmoead} introduced a Mixture-of-Experts architecture to transform single-class models into a unified model. DiAD \cite{he2024diffusion} leveraged diffusion models for multi-class AD by integrating a semantic-guided network that helps preserve semantic information in the reconstruction of a Latent Diffusion Model. More Recently, GLAD \cite{yao2024glad} proposed combining the global and local adaptive mechanisms to improve the reconstruction performance of diffusion models. However, diffusion-based methods indiscriminately apply noise to the entire image and lack an explicit mechanism for learning how to reconstruct abnormal regions from their normal surrounding regions.

\section{Preliminaries}

Denoising Diffusion Probabilistic Models (DDPM) are based on the idea of progressively perturbing data samples into noise via a forward process, and reversing this process to generate new data samples. This section introduces the fundamental concepts and notations required for understanding and applying diffusion models introduced in \cite{ho2020denoising} and \cite{songdenoising}.

\subsection{Forward Diffusion Process}

Let $\mathcal{X}=\{\xx^{(i)}\}^N_{i=1}$ denote the data samples (\ie, images), where $\xx^{(i)} \in \mathbb{R}^{H \times W \times C}$ is the \textit{i}-th image, with $H$, $W$, and $C$ representing its height, width and number of channels, respectively.  The forward Markov diffusion process consists in gradually corrupting data samples by adding Gaussian noise for $t$ time-steps where $t \in [1, T]$. Following \cite{ho2020denoising}, the forward noising process 
can be characterized as follows: 
\begin{equation}
\label{xt}
q(\xx_t|\xx_{t-1})\,=\,\mathcal{N}(\xx_t;\sqrt{1-\beta_t}\xx_t,\beta_t\mathbf{I}).
\end{equation}
Here, $\beta_i$ is a noise scheduling parameter that controls the amount of noise added at each step, such that $\xx_{T} \sim \mathcal{N}(\mathbf{0}, \mathbf{I})$ for a sufficiently large $T$. 
The marginal distribution at the moment $t$ can then be explicitly defined from $\xx_{0}$ as:
\begin{equation}
\label{xt-t0}
q(\xx_t|\xx_{0})\,=\,\mathcal{N}(\xx_t;\sqrt{\bar{\alpha}_t}\xx_0,(1-\bar{\alpha}_t)\mathbf{I}),
\end{equation}
where $\alpha_t=1-\beta_t$ and $\bar{\alpha}_t=\prod_{i=1}^T \alpha_i$.

\subsection{Reverse Diffusion Process}

In the reverse process, which iteratively removes noise through a series of denoising steps to generate new samples from Gaussian noise, the prior probability at each step can be modeled as a Gaussian distribution:
\begin{equation}
\label{prior-prob}
p(\xx_{t-1}|\xx_{t})\,=\,\mathcal{N}(\xx_{t-1};\mu_{\theta}(\xx_t,t), \beta_{t} \mathbf{I}),
\end{equation}
in which $\beta_t$ is fixed for each $t$, and the mean functions $\mu_{\theta}(\xx_t,t)$ parameterized by $\theta$ are trainable: 
\begin{equation}
\label{mu}
\mu_{\theta}(\xx_t,t)\,=\,\frac{1}{\sqrt{\bar {\alpha_t}}}\Big(\xx_t-\frac{\beta_t}{\sqrt{1-\bar{\alpha}_t}}\ee_{\theta}(\xx_t,t)\Big).
\end{equation}

This reverse process is trained with the variational lower bound of the log-likelihood of $\xx_0$, \ie: 
\begin{equation}
\begin{aligned}
 \mathcal{L}(\theta)\, = \,&-p(\xx_0|\xx_1) \\ &+ \hfill
\sum_t \mathcal{D}_{KL}\big(q^*(\xx_{t-1}|\xx_t,\xx_0)\,\|\,p_{\theta}(\xx_{t-1}|\xx_t)\big).\hfill \\
\end{aligned}
\end{equation}

By applying the reparameterization trick (\ie, parameterizing the noise function $\ee_{\theta}$ as a neural network), the model can be trained using a simple mean-square error loss between the ground truth sampled Gaussian noise and the predicted noise:
\begin{equation}
\min_\theta \ \ \mathbb{E}_{\boldsymbol{\xx}_0 \sim q\left(\boldsymbol{\xx}_0\right), \ee, t}\left[\left\|\ee-\ee_\theta\left(\boldsymbol{\xx}_t, t\right)\right\|_2^2\right],
\end{equation}
with $\ee \sim \mathcal{N}(\mathbf{0}, \mathbf{I})$. Thus, starting from a noise vector $\xx_{T} \sim \mathcal{N}(\mathbf{0}, \mathbf{I})$ sampling is achieved by iteratively applying the learned reverse process to gradually reconstruct a clean sample $\xx_{0}$.

\begin{figure*}[ht!]
\centering

\includegraphics[width=0.9\linewidth]{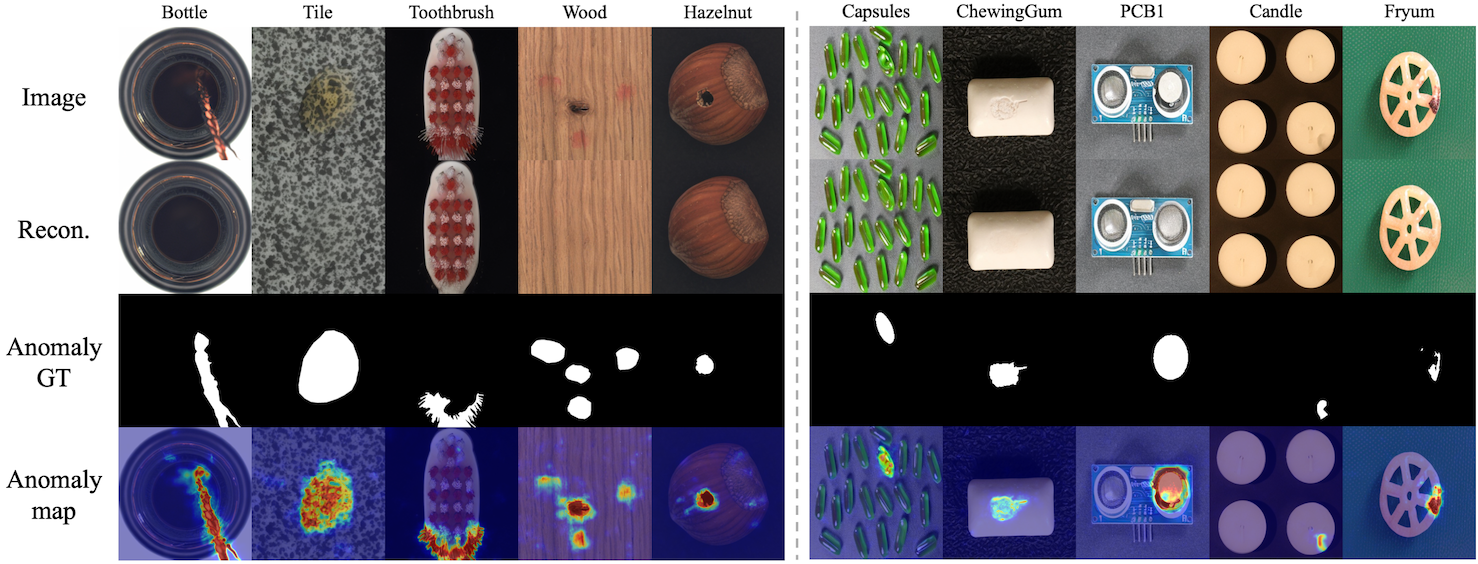}
     \caption{\textbf{Qualitative results.} From \textit{top} to \textit{bottom}: the original input image (with anomalies), \Ours{} reconstruction, the ground truth mask, and the predicted anomaly mask. Examples are depicted for two datasets (\MVTEC{} on the left side and VisA on the right side) and across multiple anomalies with diverse complexity.}
\label{fig:qualitative}
\end{figure*}

\section{Method}

We resort to diffusion models to characterize the distribution of normal images, and reconstruct the normal counterparts of target images to detect and localize anomalies. However, standard diffusion models are inherently designed to generate new samples starting from pure noise and not altering selective regions of an input image. Moreover, by applying the forward and then reverse diffusion processes to reconstruct the normal counterpart of an anomalous image, depending on time-step $t$, the information in the normal regions can be partially degraded or lost. This degradation leads to a suboptimal reconstruction of normal areas, which is not ideal for unsupervised anomaly detection as input images and their reconstructed counterparts are compared to detect abnormal regions. 

To address this issue, we propose a modified diffusion model (\Ours) that selectively alters only the anomalous regions of an image, while leaving the normal areas intact. This approach allows for preserving the normal parts of anomalous images while altering abnormal regions, based on the learned distribution of normal images. In this section, we detail the different components of our approach, whose overall pipeline is depicted in \cref{fig:method}.

\subsection{Modeling anomalies as noise in latent space}

We employ a pre-trained Variational Auto-Encoder (VAE) \cite{rombach2022high} to project images into a latent space where the diffusion process is performed. We chose to work in this latent space for five key reasons: \textit{i)} anomalies in the image can be effectively interpreted as noise within the latent space, which aligns well with the operational framework of diffusion models; it also \textit{ii)} alleviates the problem of ``identity shortcut'' which causes anomalies to be preserved in the reconstruction; \textit{iii)} increases computational efficiency; \textit{iv)} enhances stability, especially when training data is limited; and \textit{v)} improves the quality of generated samples.

Denoting the encoder and decoder networks of the VAE as $\phi_{E}$ and $\phi_{D}$, an input image $\xx^{(i)}$ is projected into a latent space representation $\zz^{(i)}=\phi_{E}(\xx^{(i)})$, where $\zz^{(i)} \in \mathbb{R}^{H' \times W' \times C'}$. It is important to note that, due to the unsupervised nature of this method, we only have access to normal images during the training phase.

\noindent \textit{Limitations of standard diffusion models.} There are certain limitations to adding noise to the whole input image (or its latent representation) as in standard diffusion models. First, the added noise also affects normal regions. Consequently, during the forward diffusion process, normal regions may experience a partial or complete loss of information, making it difficult to fully recover the input image and potentially leading to their misinterpretation as anomalous areas when compared to the input image. Furthermore, abnormal regions should be reconstructed with respect to the surrounding normal areas. However, if normal patches are altered due to the forward diffusion process, they cannot be fully exploited to reconstruct abnormal patches, a critical step for anomaly detection.

To alleviate the aforementioned limitations, we propose to integrate random masking into the forward diffusion process. Given the latent features of an input normal sample $\zz_0 \sim p\left(\zz_0\right)$, we spatially divide $\zz_0$ into non-overlapping noisy $\zz_0^n$ (non-masked) and visible $\zz_0^v$ patches (masked) using a random mask with a random masking ratio ($r_{mask}$). Afterward, during the forward process, only the noisy patches in $\zz_0^n$ are gradually diffused, while the visible patches $\zz_0^v$ remain unchanged:
\begin{equation}
q(\zz_t^k|\zz_{0}^k) \, = \,
\left\{ 
  \begin{array}{l}
    \mathcal{N}(\zz_t^{n};\sqrt{\bar{\alpha}_t}\zz_0^{n},(1-\bar{\alpha}_t)\mathbf{I}), \ \ \textsl{if $k=n$}  \\[3pt]
    \zz_0^v, \ \ \textsl{otherwise.}       \\
  \end{array}
\right.
\label{eq:masking}
\end{equation}

This way, visible patches in the image can represent normal regions that have been unaltered during the forward process, while noisy patches could mimic anomalous areas.

\subsection{Deviation instead of Diffusion}
While we motivated the use of a masking strategy (\cref{eq:masking}) during the forward diffusion process, conditioning the reverse process and training objective on the masks used for the forward pass is suboptimal and impractical. First, accessing to prior knowledge on the location of abnormal regions during test time is unrealistic. Secondly, the denoising operation in \cref{prior-prob} is not applicable to \textit{non-noisy} patches. Thus, we need to integrate a mechanism that optimizes the training objective and applies the reverse process to both noisy and visible patches.

To achieve this, we reformulate the standard diffusion model to better suit our needs. Let us first consider the noisy patches. Based on \cref{xt}, for noisy patches in the latent space we have:
\begin{equation}
\zz_{t}^{n} \, = \, \sqrt{1-\beta_{t}}\zz_{t-1}^{n} + \sqrt{\beta_{t}}\ee
\end{equation}
Based on \cref{xt-t0}, $\zz_{t}^{n}$ could be explicitly obtained with
\begin{equation}
\label{z_t}
\zz_{t}^{n} = \sqrt{\bar{\alpha_{t}}}\zz_{0}^{n} + \sqrt{1-\bar{\alpha_{t}}}\ee
\end{equation}
We can then rewrite \cref{z_t} as:
\begin{equation}
\begin{aligned}
\zz_{t}^{n} & \, = \,  (1-(1-\sqrt{\bar{\alpha_{t}}}))\zz_{0}^{n} + \sqrt{1-\bar{\alpha_{t}}}\ee \hfill \\[.3em]
& \, = \, \zz_{0}^{n} + \sqrt{1-\bar{\alpha_{t}}}\ee -(1-\sqrt{\bar{\alpha_{t}}})\zz_{0}^{n} \\[.3em]
& \, =\,  \zz_{0}^{n} + \underbrace{\sqrt{1-\bar{\alpha_{t}}}\left(\ee - \frac{1-\sqrt{\bar{\alpha_{t}}}}{\sqrt{1-\bar{\alpha_{t}}}}\zz_{0}^{n}\right)}_{\textit{``Deviation from Normality''}} \\
\end{aligned}
\end{equation}
In the above equation, the second term measures the deviation of $\zz_{t}^{n}$ from $\zz_{0}^{n}$, which represents a normal image. In summary, we have that:
\begin{equation}
\begin{aligned}
\zz_{t}^{n} & \, =\,  \zz_{0}^{n} + \sqrt{1-\bar{\alpha_{t}}}\etabold \\
\etabold & \, = \,\left(\ee - \frac{1-\sqrt{\bar{\alpha_{t}}}}{\sqrt{1-\bar{\alpha_{t}}}}\zz_{0}^{n}\right),
\end{aligned}
\end{equation}
where $\etabold$ is a term based on the time-step $t$, $\zz_{0}$, and $\ee$, which points to the ``\emph{Direction of Deviation}'' (DoD). One interesting characteristic of this approach is that the patches will remain untouched in the forward process if $\etabold=\mathbf{0}$. 

The reverse process aims to correct the deviation from a normal image by progressively removing noise through a learned denoising function $p(\zz_{t-1}|\zz_{t})$. This could be achieved by training a neural network to closely approximate the true reverse path by predicting the DoD at each time-step. The main objective can thus be defined as:
\begin{equation}
\min _\theta \ \ \mathbb{E}_{\boldsymbol{\zz}_0 \sim q\left(\boldsymbol{\zz}_0\right), \ee, t}\left[\left\|\etabold-\etabold_\theta\left(\boldsymbol{\zz}_t, t\right)\right\|_2^2\right].
\end{equation}

This approach encourages the model to predict the DoD from normality at each time-step, facilitating accurate correction in the reverse process. As a result, our DoD predictor ($\etabold_\theta$) learns a robust denoising trajectory that can correct the noisy, abnormal patches with respect to the surrounding normal ones, while keeping the normal patches untouched by simply predicting zero as their DoD. Furthermore, to expose the model to more structured anomalies during training, we also propose to incorporate a strategy based on \textit{patch shuffling}. Since the latent space follows a normal distribution, for a portion of patches (specified with $r_{shuffle}$), instead of adding Gaussian noise, we can replace them with other patches taken from images in the same batch. This strategy could ultimately result in better localization of large and structured anomalies.

\begin{table*}[ht!]
    \centering
    \footnotesize
    \caption{\textbf{Quantitatve evaluation  on \textbf{MVTec-AD}.} Image and Pixel-level results 
    on \textit{multi-class} anomaly detection. 
    Best method (per metric) is highlighted in \first{blue}, whereas \second{red} is used to denote the second-best approach. Differences with the second (or best) approach in (\imp{gray}).}
    \label{tab:both-mvtec}
    \begin{tabular}{l | ccc | cccc}
 \multicolumn{1}{c|}{\multirow[b]{2}{*}{Method}} & \multicolumn{3}{c|}{Image-level} & \multicolumn{4}{c}{Pixel-level}\\
 & \footnotesize{AUROC} & \footnotesize{AUPRC} & \footnotesize{f1$_{\text{max}}$} &  \footnotesize{AUROC} & \footnotesize{AUPRC} & \footnotesize{f1$_{\text{max}}$} & \footnotesize{AUPRO} \\
 \toprule

RD4AD \tiny\textsl{CVPR'22} & 94.6 & 96.5 & 95.2 & 96.1 & 48.6 & 53.8 & 91.1 \\ 
UniAD \tiny\textsl{NeurIPS'22} & 96.5 & 98.8 & 96.2 & 96.8 & 43.4 & 49.5 & 90.7 \\
SimpleNet \tiny\textsl{CVPR'23} & 95.3 & 98.4 & 95.8 & 96.9 & 45.9 & 49.7 & 86.5 \\
DeSTSeg \tiny\textsl{CVPR'23} & 89.2 & 95.5 & 91.6 & 93.1 & 54.3 & 50.9 & 64.8 \\
DiAD \tiny\textsl{AAAI'24} & 97.2 & 99.0 & 96.5 & 96.8 & 52.6 & 55.5 & 90.7\\
MoEAD \tiny\textsl{ECCV'24} & 98.0 & 99.3 & 97.5 & 96.9 & 49.8 & 43.5 & 91.0 \\
GLAD  \tiny\textsl{ECCV'24} & 97.5 & 99.1 & 96.6 & 97.4 & \second{60.8} & \second{60.7} & 93.0\\
MambaAD \tiny\textsl{NeurIPS'24} & \second{98.6} & \second{99.6} & \second{97.8} & \second{97.7} & 56.3 & 59.2 & \second{93.1}\\
\midrule
\Ours{} (\textit{Ours}) & \first{99.3}\imp{+0.7} & \first{99.8}\imp{+0.2} & \first{98.5}\imp{+0.7} & \first{98.4}\imp{+0.7} & \first{74.9}\imp{+14.1} & \first{69.7}\imp{+9.0} & \first{94.9}\imp{+1.8} \\
\bottomrule
\end{tabular}

\end{table*}

\subsection{Anomaly detection} 

In order to detect anomalies at inference time, we first need to adapt the sampling of DDPM  (\cref{prior-prob}) to fit the proposed configuration for correcting the deviation from normality. However, before describing this step, it is worth mentioning that sampling with DDPMs requires many reverse sampling steps until $t=0$, leading to a significant computational burden. In addition, DDPMs introduce noise over the whole image at each step $t$, which goes against our objective of keeping normal patches unchanged. Consequently, we resort to the Denoising Diffusion Implicit Model (DDIM) \cite{songdenoising} during inference, which modifies the sampling process by making it deterministic instead of stochastic. To correct the deviation using DDIM sampling, with a trained model $\etabold_{\theta}$, the predicted normal image ($\tilde{\zz}_{0}$) at each time-step $t$ is:
\begin{equation}
\tilde{\zz}_{0} \, = \, \zz_{t} - \sqrt{1-\bar{\alpha_{t}}}\etabold_{\theta}(\zz_{t}, t).
\end{equation}
Therefore, the reverse process becomes:
\begin{equation}
\tilde{\zz}_{t-1} \, = \, \tilde{\zz}_{0} \, + \sqrt{1-\bar{\alpha}_{t-1}}\!\!\!\!\!\!\!\!\underbrace{\etabold_{\theta}(\zz_{t}, t)}_{\textrm{``\textit{Direction of deviation}"}}
\label{eq:dirdev}
\end{equation}

In the above equation, $\tilde{\zz}_{0}$ is the predicted $\zz_{0}$, and $\etabold_{\theta}(\zz_{t}, t)$ the DoD. Note that \cref{eq:dirdev} 
enables a deterministic trajectory from 
$\zz_{\tau}$ to $\zz_{0}$, which accommodates our setting since it avoids introducing any noise. It is noteworthy to mention that a key advantage of DDIM over DDPM is that, by eliminating the noise at each step and making the denoising process more direct, high-quality samples can be generated in fewer steps. This results in a faster convergence to normal samples.

Intuitively, as our \Ours{}  model has been solely trained using normal images, when an anomalous image is provided the anomalous regions will fall outside the learned distribution. Consequently, the model will consider them as non-masked noisy patches, which will be denoised during the reverse process. On the other hand, if the patches do not correspond to anomalous regions, the model will consider them as masked normal patches and they will remain unchanged during the reconstruction. Furthermore, in contrast to standard diffusion processes that add noise to the whole image, the proposed model allows leveraging surrounding normal information when reconstructing abnormal regions.

\mypar{Leveraging multi-scale information.} Pixel-level reconstruction discrepancy is widely used in reconstruction-based UAD approaches mainly due to this ability to produce detailed anomaly maps. Nevertheless, some anomalous regions might not be revealed using this approach, for example, if they only present small differences in color compared to the normal ones. On the other hand, reconstruction discrepancy measured in the latent space can capture more subtle anomalies, however, it yields coarser anomaly maps due to the reduced spatial dimension of the latent space. 

Based on these observations, we propose a multi-scale strategy where discrepancies at both pixel-space and latent-space are leveraged jointly to localize potential anomalies. More concretely, in addition to the reconstructed ``normal'' feature embedding $\tilde{\zz}_{0}$, we also generate a reconstructed normal sample in the image-space as $\tilde{\xx}_{0} = \phi_{D}(\tilde{\zz}_{0})$, where $\phi_{D}$ is the pre-trained VAE decoder. The predicted anomaly map can be thus obtained using the geometric mean of latent-level and pixel-level discrepancies as:
\begin{equation}
    \aaa \, = \,  \sqrt{\frac{1}{\gamma_{l}.\gamma_{p}}\min(\|\tilde{\zz}_{0} - \zz_{0}\|, \gamma_{l}) \cdot  \min(\|\tilde{\xx}_{0} - \xx_{0}\|, \gamma_{p})}.
    \label{eq:multilevel}
\end{equation}
Here, $\gamma_{l}$ and $\gamma_{p}$ respectively denote the latent-level and pixel-level thresholds, which prevent assigning excessive weight to patches with significant deviations in either the image or the latent space. For instance, if a black patch is reconstructed as white, this does not necessarily indicate that the patch is more anomalous compared to a patch reconstructed as red.

\section{Experiments}

\begin{table*}[ht!]
    \centering
    \footnotesize
    \caption{\textbf{Quantitatve evaluation  on \textbf{VisA}.} Image and Pixel-level results 
    on \textit{multi-class} anomaly detection. 
    The best method (per metric) is highlighted in \first{blue}, whereas \second{red} is used to denote the second-best approach. Differences with second (or best) approach in (\imp{gray}).}
    \label{tab:both-visa}
    \begin{tabular}{l | ccc | cccc}
 \multicolumn{1}{c|}{\multirow{2}{*}{Method}} & \multicolumn{3}{c|}{Image-level} & \multicolumn{4}{c}{Pixel-level}\\
  & \footnotesize{AUROC} & \footnotesize{AUPRC} & \footnotesize{f1$_{\text{max}}$} &  \footnotesize{AUROC} & \footnotesize{AUPRC} & \footnotesize{f1$_{\text{max}}$} & \footnotesize{AUPRO} \\
 \toprule

RD4AD \tiny\textsl{CVPR'22} & 92.4 & 92.4 & 89.6 & 98.1 & 38.0 & 42.6 & 91.8 \\ 
UniAD \tiny\textsl{NeurIPS'22} & 88.8 & 90.8 & 85.8 & 98.3 & 33.7 & 39.0 & 85.5 \\
SimpleNet \tiny\textsl{CVPR'23} & 87.2 & 87.0 & 81.8 & 96.8 & 34.7 & 37.8 & 81.4 \\
DeSTSeg \tiny\textsl{CVPR'23} & 88.9 & 89.0 & 85.2 & 96.1 & \second{39.6} & 43.4 & 67.4 \\
DiAD \tiny\textsl{AAAI'24} & 86.8 & 88.3 & 85.1 & 96.0 & 26.1 & 33.0 & 75.2 \\
MoEAD \tiny\textsl{ECCV'24} & 93.0 & \second{95.1} & \second{89.8} & \first{98.7} & 36.6 & 41.0 & 88.6 \\
GLAD  \tiny\textsl{ECCV'24} & 91.8 & 92.9 & 88.0 & 97.4 & 34.6 & 40.3 & \second{92.0} \\
MambaAD \tiny\textsl{NeurIPS'24} & \second{94.3} & 94.5 & 89.4 & 98.5 & 39.4 & \second{44.0} & 91.0\\
\midrule
\Ours{} (\textit{Ours}) & \first{96.4}\imp{+2.1} & \first{96.8}\imp{+1.7} & \first{92.2}\imp{2.4} & \second{98.5}\imp{-0.2} & \first{51.3}\imp{+11.7} & \first{51.2}\imp{+7.2} & \first{92.1}\imp{+0.1} \\

\bottomrule
\end{tabular}
\end{table*}

\subsection{Setting}

\mypar{Datasets.} We evaluate our method on two well-known anomaly detection datasets. \textbf{MVTec-AD~\cite{bergmann2019mvtec}}  simulates real-world industrial production scenarios, filling the gap in unsupervised anomaly detection. It consists of 5 types of textures and 10 types of objects, in 5,354 high-resolution images from different domains. \textbf{VisA~\cite{zou2022spot}} consists of 10,821 high-resolution images, with 78 types of anomalies. It comprises 12 subsets corresponding to distinct objects categorized into three object types: Complex structure, Multiple instances, and Single instance. Note that ablations are performed on \MVTEC{}. 
More information about the datasets is provided in \appensecref{sec:infodata}.

\mypar{Evaluation Metrics.} Following prior literature \cite{he2024diffusion,he2024mambaad}, we use Area Under the Receiver Operating Characteristic Curve (AUROC), Area Under Precision-Recall Curve (AUPRC\footnotemark), and f1-score-max (f1$_{\text{max}}$) to measure performance in both anomaly detection (i.e., \textit{image-level}) and anomaly localization (i.e., \textit{pixel-level}). Furthermore, Area Under Per-Region-Overlap (AUPRO) is employed for pixel-level anomaly localization.

\footnotetext{Also called Average Precision(AP) or AUPR in the literature}

\mypar{Baselines.} We compare our approach to eight recently-proposed approaches for multi-class UAD: RD4AD \cite{deng2022anomaly}, UniAD \cite{you2022unified}, SimpleNet \cite{liu2023simplenet}, DeSTSeg \cite{zhang2023destseg}, DiAD \cite{he2024diffusion}, MoEAD\cite{mengmoead}, GLAD \cite{yao2024glad}, and MambaAD \cite{he2024mambaad}.

\mypar{Implementation Details.}
To map data into the latent space using a pre-trained VAE, we used a pre-trained perceptual compression model \cite{rombach2022high}, which consists of an auto-encoder trained by a combination of a perceptual loss \cite{zhang2018unreasonable} and a patch-based adversarial objective that down-samples the image by a factor 8. Similar to VAEs\cite{kingma2013auto}, a regularization loss measuring the KL divergence between latent features and $\mathcal{N}(\mathbf{0}, \mathbf{I})$ is used to avoid the latent space having an arbitrarily high variance. 

For our model $\etabold_\theta$ predicting the \emph{direction of deviation} (DoD), following \cite{rombach2022high}, we employed an attention UNet architecture with timestep embedding and a squared-cosine-beta scheduler \cite{nichol2021improved}. At each iteration and for each training sample, we first randomly select a timestep $t$ from a uniform distribution between 1 and $T$ (where $T$=10). Then, we sample a masking ratio ($r_{\text{mask}}$) from a uniform distribution $U[0, 0.7]$ and a latent patch size from $\{1,2,4,8\}$ corresponding to pixel-wise patch sizes of $\{8\!\times\!8, 16\!\times\!16, 32\!\times\!32, 64\!\times\!64\}$. Using these sampled values, we generate a random mask with the specified ratio. Moreover, in our implementation, the replacement ratio ($r_{shuffle}$) is sampled in each iteration from $U[0, 0.3]$. Finally, $\zz_{t}$ is obtained using \cref{eq:masking}, and serves as input to train our model. This model was trained 800 epochs for experiments on the MVTec dataset and 200 epochs for VisA, in both cases using a batch size of 128 and a single A6000 GPU. We used the AdamW optimizer with a Cosine Annealing scheduler with warm-up, with an initial learning rate set to $10^{-4}$, decaying to a minimum of $10^{-5}$.  The VAE remained frozen throughout training.

\subsection{Results}

\subsubsection{Quantitative Results} \textbf{\textit{\MVTEC{}:}} 
\cref{tab:both-mvtec} reports the performance for all methods, where we can see that the proposed \Ours{} obtains the overall best performance across all metrics for both image-wise and pixel-wise multi-class anomaly detection. 
More concretely, \Ours{} achieves 0.7\%, 0.2\%, 0.7\% improvements over the second best approach, i.e., MambaAD \cite{he2024mambaad}, for image-level AUROC, AUPRC, and f1$_{\text{max}}$, respectively. The boost in performance is even more notable at pixel-level, 
an arguably more complex scenario. In particular, our method yields improvements of 0.2\%, 14.1\%, 9.0\%, and 1.8\% over the existing state-of-the-art multi-class UAD models measured by pixel-level AUROC, AUPRC, f1$_{\text{max}}$, and AUPRO, respectively. \textbf{\textit{VisA:}} Results in \cref{tab:both-visa} confirm the trend observed in \MVTEC{}, with our approach outperforming recent methods in both pixel-level and image-level metrics. Particularly, \Ours{} obtains around 2\% performance gain over the second-best approach in image-level metrics, and exhibits significant pixel-level improvement over all methods in terms of AUPRC, f1$_{\text{max}}$, and AUPRO, yet ranking second for pixel-level AUROC. Detailed per-category results are provided in \appensecref{sec:per-category}.

\subsubsection{Qualitative Results}
\begin{figure}[!t]
\centering
\includegraphics[width=\linewidth]{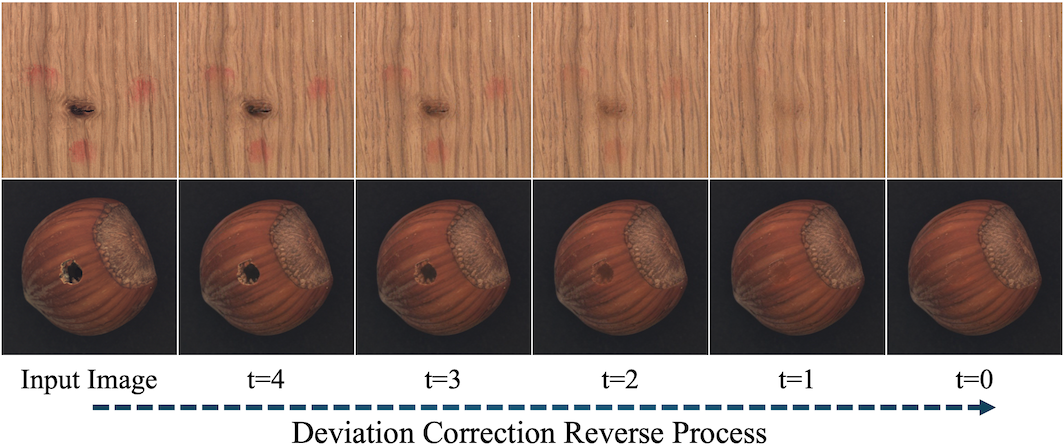}
     \caption{\textbf{Visualization of deviation correction through the reverse process steps.} Abnormal areas are progressively corrected, while fine normal details are preserved during the reverse process.}
\label{fig:reverse}
\end{figure}

The quantitative results presented above are supported by visual examples showcasing the effectiveness of our approach in identifying anomalous image regions. In \cref{fig:qualitative}, we depict anomalous samples from both datasets  
and their reconstruction using our model. As it can be observed, abnormal regions 
are properly modified during the deviation correction process and replaced by their normal counterpart. In contrast, the normal regions remained unchanged, our model preserving the fine details in these regions. This enables a precise detection and localization of abnormalities when contrasting the input image to its reconstructed version (\cref{fig:qualitative}, \textit{last row}). Moreover, \cref{fig:reverse} visually depicts 
the deviation correction reverse process from \Ours{}, which showcases how abnormal areas are progressively removed, until a \textit{cleaned} version of the input image is generated.

\subsubsection{Ablation Studies}

\mypar{Effect of backward deviation correction strategy and number of time-steps.} 
In this section, we explore different 
strategies for the deviation correction process during inference. 
In \cref{eq:dirdev}, we explained how the proposed deviation correction formulation can be used in the DDIM reverse sampling 
to progressively correct the abnormalities in each time-step $t$. An alternative strategy would be to directly replace $\zz_{t-1}$ with $\tilde{\zz}_{0}$ at each time-step. This approach allows the model to correct the input image based on its initial prediction at the first time-step, and then continue refining any remaining abnormalities in subsequent steps. 

We evaluated the performance for different number of reverse steps 
using both approaches, where the first approach is referred to as $\tilde{\zz}_{t-1}$ and the latter as $\tilde{\zz}_{0}$ in  \cref{tab:ablation_ddim_timesteps}. Note that, for a single reverse step, both approaches are identical and yield the same results. As reported in \cref{tab:ablation_ddim_timesteps}, the first strategy ($\tilde{\zz}_{t-1}$) performs better in image-level metrics, while the second yields slightly better performance 
in terms of pixel-level metrics. Furthermore, the performance is generally consistent as the number of reverse time-steps increases, making a lower number of time-steps (but larger than 1) more favorable due to its decreased computational cost.

\begin{table}[!ht]
    \centering
    \caption{Impact of the correction strategy at each time-step and number of reverse steps.} 
    \footnotesize
    \label{tab:ablation_ddim_timesteps}
    \begin{tabular}{@{}c@{\hskip 4pt}|@{\hskip 4pt}c@{\hskip 4pt}|@{\hskip 4pt}c@{\hskip 4pt}|@{\hskip 4pt}c@{}}
 \small{Correction} & \small{Reverse} & \small{Image-level} & \small{Pixel-level}\\
 \small{strategy} & \small{steps} & \tiny{AUROC\,/AUPRC\,/f1$_{\text{mask}}$} & \tiny{AUROC\,/AUPRC\,/f1$_{\text{mask}}$\,/AUPRO} \\
 \toprule
 
\small{$\tilde{\zz}_{0}$} & 1 & \imgperf{98.4}{99.3}{97.9} & \pixperf{98.4}{72.9}{68.8}{94.3} \\
\small{$\tilde{\zz}_{0}$} & 2 & \imgperf{99.2}{99.7}{98.1} & \pixperf{\textbf{98.4}}{\textbf{75.1}}{\textbf{69.9}}{94.7} \\
\small{$\tilde{\zz}_{0}$} & 5 & \imgperf{99.2}{99.7}{97.9} & \pixperf{98.2}{74.3}{69.3}{94.6} \\
\small{$\tilde{\zz}_{0}$} & 10 & \imgperf{98.9}{99.5}{97.6} & \pixperf{97.8}{73.6}{68.8}{94.2} \\
\midrule
\small{$\tilde{\zz}_{t-1}$} & 2 & \imgperf{99.3}{99.8}{98.3} & \pixperf{98.4}{75.0}{69.8}{94.4} \\
\small{$\tilde{\zz}_{t-1}$} & 5 & \imgperf{\textbf{99.3}}{\textbf{99.8}}{\textbf{98.5}} & \pixperf{98.4}{74.9}{69.7}{\textbf{94.9}} \\
\small{$\tilde{\zz}_{t-1}$} & 10 & \imgperf{99.3}{99.8}{98.5} & \pixperf{98.4}{74.9}{69.8}{95.0} \\

\bottomrule
\end{tabular}
\end{table}

\mypar{Effect of leveraging discrepancies at pixel, latent, and visual feature levels.}
In reconstruction-based anomaly detection, mapping the differences between inputs and their reconstructions to anomaly maps is essential. Thus, we now investigate several alternatives to the proposed strategy, presented in \cref{eq:multilevel}. First, we consider using only the difference in pixel values, i.e., $\Delta\xx=|\xx_{T} - \xx_{0}|$, which are clipped and scaled by $\gamma_l=0.4$, as: $\frac{1}{\gamma_l} \min(\Delta\xx,\gamma_l)$. 
However, this option may not be optimal, as anomalies might appear with the same color as normal regions. Then, we resort to the difference in the latent space ($\Delta\zz=|\zz_{T} - \zz_{0}|$), which is similarly clipped and scaled by the same value, and then resized to match the target image size. Additionally, we explored the impact of the arithmetic mean: $\frac{1}{2\gamma_l} \min(\Delta\zz,\gamma_l)+\frac{1}{2\gamma_p} \min(\Delta\xx,\gamma_p)$. 
Finally, we extracted the visual features from a pre-trained ResNet50 \cite{he2016deep} for both the input images and their reconstructions and compared them using the cosine similarity to localize anomalies. The results for anomaly detection and localization using different levels of discrepancies
are presented in \cref{tab:ablation_dissimlaritt}. These results support our approach (i.e., Geometric mean, \cref{eq:multilevel}), and demonstrate that combining pixel- and latent-space discrepancies with the geometric mean yields superior performance to considering these discrepancies separately.

\begin{table}[h]
    \centering
    \footnotesize
    \caption{\textbf{Different levels of discrepancy.} Results on \MVTEC{}.} 
    \label{tab:ablation_dissimlaritt}
    \begin{tabular}{@{}l@{\hskip 3pt}|@{\hskip 3pt}c@{\hskip 3pt}|@{\hskip 3pt}c@{}}
 \small{\multirow{2}{*}{Dissimilarity}} & \small{Image-level} & \small{Pixel-level}\\
    & \tiny{AUROC\,/AUPRC\,/f1$_{\text{mask}}$} & \tiny{AUROC\,/AUPRC\,/f1$_{\text{mask}}$\,/AUPRO} \\
 \toprule
 
 \small{Pixel diff. ($\Delta\xx$)}& \imgperf{98.9}{99.6}{97.7} & \pixperf{97.5}{68.0}{63.7}{92.1} \\
 \small{Latent diff. ($\Delta\zz$)} & \imgperf{99.2}{99.7}{98.3} & \pixperf{98.2}{70.5}{67.5}{93.5}\\
 \small{Arithmetic ($\Delta\xx$, $\Delta\zz$)} &  \imgperf{\textbf{99.3}}{\textbf{99.8}}{\textbf{98.5}} & \pixperf{98.2}{73.5}{68.8}{94.1}\\
\small{Geometric ($\Delta\xx$, $\Delta\zz$)} & \imgperf{\textbf{99.3}}{\textbf{99.8}}{\textbf{98.5}} & \pixperf{\textbf{98.4}}{\textbf{74.9}}{\textbf{69.7}}{\textbf{94.9}}\\
 \small{Features (ResNet50)} & \imgperf{98.5}{99.3}{97.3} & \pixperf{97.4}{66.7}{64.6}{91.7}\\
 
\bottomrule
\end{tabular}
\end{table}

\section{Conclusions}

We proposed a reformulation of standard diffusion models, specifically designed to modify abnormal regions in target images without affecting normal areas. By integrating masked strategy to the new formulation in the latent space, \Ours{} progressively detects and corrects the deviations from normality, enabling precise localization of abnormalities. Our comprehensive Quantitative and Qualitative results demonstrate the superiority of our model compared to state-of-art UAD methods in the unified multi-class context.




\bibliographystyle{ieeenat_fullname}
\bibliography{main}

\clearpage

\appendix
\clearpage
\clearpage
\setcounter{page}{1}
\setcounter{table}{0}
\setcounter{figure}{0}
\setcounter{section}{0}
\maketitlesupplementary

\section{Additional information about the datasets}
\label{sec:infodata}

\textbf{MVTec-AD~\cite{bergmann2019mvtec}}: The training set contains 3,629 images with only anomaly-free samples. The test set consists of 1,725 images, comprising 467 normal samples and 1,258 abnormal ones. The anomalous samples exhibit diverse defects, including surface imperfections (e.g., scratches, dents), structural anomalies such as deformed object parts, and defects characterized by missing object components. Pixel-level annotations are provided for the anomaly localization evaluation.  

\noindent\textbf{VisA~\cite{zou2022spot}}:  Including 9,621 normal images and 1,200 anomaly images with 78 types of anomalies. The VisA dataset comprises 12 subsets, each corresponding to a distinct object. 12 objects could be categorized into three object types: Complex structure, Multiple instances, and Single instance. The anomalous images exhibit a range of flaws, including surface defects such as scratches, dents, color spots, and cracks, as well as structural defects like misalignments or missing components.

\noindent\textbf{Additional datasets}: In addition to the \MVTEC{}  and Visa datasets, we have used two additional multi-class anomaly detection datasets to evaluate the generalizability of our model: the Metal Parts Defect Detection (\textbf{MPDD} \cite{jezek2021deep}) dataset and the Real-Industrial Anomaly Detection (\textbf{Real-IAD} \cite{wang2024real}) dataset. The MPDD dataset contains 1,346 multi-view images with pixel-precise defect annotations for \underline{6} distinct industrial metal products. Real-IAD presents a more challenging scenario, encompassing objects from  \underline{30} categories with a total of 150K high-resolution, multi-view images, including 99,721 normal images and 51,329 anomalous images. The anomalies in Real-IAD span a broad spectrum, including pits, deformations, abrasions, scratches, damages, missing parts, foreign objects, and contamination.

\section{Detailed per-category results}
\label{sec:per-category}
Detailed image-level and pixel-level Mc-UAD anomaly detection results for all categories of \MVTEC{} of the proposed method, as well as comparing methods are presented in \cref{tab:img-mvtec} and \cref{tab:px-mvtec}, respectively. Similarly, detailed per-category results for the VisA dataset are presented in \cref{tab:img-visa} and \cref{tab:px-visa}. These results highlight the effectiveness of our approach, demonstrating its superiority over various state-of-the-art (SoTA) methods across most of the object categories.

\section{Quantitaive results on additional datasets}
\label{sec:quantitative-additional}
To further validate the effectiveness of the proposed method on Mc-UAD we have conducted additional experiments on new datasets, i.e. MPDD and Real-IAD. For the MPDD dataset, as there are less data samples and categories we have use Medium size model, and for Real-IAD dataset, which includes a greater number of categories with higher complexity, we utilized the X-Large model size. As depicted in \cref{tab:additional-datasets}\footnote{These results are either drawn from the original paper or papers referenced to them}, the proposed \Ours{} demonstrates superior performance, achieving improved results across both image-level and pixel-level metrics for both additional datasets.

\section{Additional ablations}
\label{sec:additional-ablation}

\subsection{Architecture design of the model.}
In this section, we investigate the impact of model size on anomaly detection and localization performance. 
As mentioned before, we have employed a UNet with attention as the backbone for our deviation correction model $\etabold_\theta$. By default, we used a Large (L) configuration with 256 channels, attention resolution of \{4,2,1\}, channel-multiplication factor of \{1,2,4\} across consecutive scales, dropout of 0.4, and 2 residual blocks. We also define model sizes XS, S, M, and XL to have 64, 128, 192, and 320 channels respectively. As indicated in \cref{tab:ablation_model_size}, the larger model sizes usually yield better image-level performance, while pixel-level performance seems to be consistent for all model sizes greater than XS.

\begin{table}[h!]
    \centering
    \footnotesize
    \caption{\textbf{Ablation studies on the impact of different model sizes.}} 
    \label{tab:ablation_model_size}
    \begin{tabular}{@{}l@{\hskip 3pt}|@{\hskip 3pt}c@{\hskip 3pt}|@{\hskip 3pt}c@{\hskip 3pt}|@{\hskip 3pt}c@{}}
 \multirow{2}{*}{\small{Model size}} & \multirow{2}{*}{\small{ \# params}} & \small{Image-level} & \small{Pixel-level}\\
 
 & & \tiny{AUROC\,/\,AUPRC\,/\,f1$_{\text{mask}}$} & \tiny{AUROC\,/\,AUPRC\,/\,f1$_{\text{mask}}$\,/\,AUPRO} \\

 \toprule
 \small{UNet-XS} & \small{25M} & \imgperf{98.1}{99.3}{97.9} & \pixperf{98.6}{73.7}{68.6}{93.5} \\
 \small{UNet-S}  & \small{99M} &  \imgperf{98.7}{99.5}{98.3} & \pixperf{\textbf{98.7}}{\textbf{75.8}}{\textbf{70.5}}{94.5}\\
 \small{UNet-M}  & \small{222M} & \imgperf{98.8}{99.6}{98.4} & \pixperf{98.6}{75.5}{70.3}{94.7}\\
 \small{UNet-L}  & \small{395M} & \imgperf{99.3}{\textbf{99.8}}{98.5} & \pixperf{98.4}{74.9}{69.7}{\textbf{94.9}}\\
 \small{UNet-XL} & \small{614M} & \imgperf{\textbf{99.3}}{99.7}{\textbf{98.7}} & \pixperf{98.5}{75.5}{70.2}{94.5}\\
\bottomrule
\end{tabular}
\end{table}

\subsection{Impact of different hyper-parameters.}
In this section, the impact of five different hyper-parameters on the Mc-UAD performance of the proposed method has been evaluated, and their results are depicted in \cref{tab:ablation_hyperparameter}. 

\noindent\textbf{\textit{Patch size}}: During each iteration of the training phase, a patch-size is randomly selected to create a random mask and reshuffle the patches. We experimented with three different sets of patch-sizes, each containing 3, 4, and 5 scales, where the scales increase exponentially (powers of 2) starting from a patch size of 1. As detailed in \cref{sec:additional-ablation}, using a larger set of patch sizes, leads to improved performance. 

\noindent\textbf{\textit{Masking ratio ($r_{\textit{mask}}$)}}: In order to  create a random mask for each input image, the ratio of masking ($r_{\textit{mask}}$) is sampled uniformly from $U[0, R_{\textit{mask}}]$. As depicted in \cref{tab:ablation_hyperparameter}, we have explored the effect of masking ratio, with three different uniform ranges with the maximum masking ratio ($R_{\textit{mask}}$) set to 0.3, 0.5, and 0.7. The results indicated that when the model is exposed to higher masking ratios during training, it could better identify out-of-distribution regions, and therefore achieve better image-level and pixel-level performance. 

\noindent\textbf{\textit{Shuffled patches ratio}}: We have investigated the effect of incorporating reshuffled patches within the batch as noise to expose the model to more structured deviations. It should be mentioned that this ratio, is the ratio of shuffled patches to non-masked patches (specified by $r_{shuffle}$), and not the whole patches. As reflected in the \cref{tab:ablation_hyperparameter}, introducing reshuffled patches as noise at a low ratio results in a slight improvement in image-level metrics, albeit at the cost of a decline in pixel-level metrics. On the other hand, replacing a high portion of noise with shuffled patches could decrease both image-level and pixel-level metrics. 

\noindent\textbf{\textit{$\gamma_{p}$ and $\gamma_{l}$}}:
$\gamma_{p}$ and $\gamma_{l}$ are devised to prevent assigning excessive weights to locations with large pixel-level and latent-level deviations. We have assessed the impact of these thresholds and the results are shown in \cref{tab:ablation_hyperparameter}. It is worth mentioning that \xmark{} in the table indicates no threshold or scaling is applied to the deviations. As it can reasonably be anticipated, the image-level metrics did not have much variation. On the other hand, results indicate that applying these thresholds could improve pixel-level metrics, where $\gamma_{p}=0.4$ and $\gamma_{l}=0.2$ yield the best results.

\begin{table}[ht!]
    \centering
    \footnotesize
    \caption{\textbf{Ablation studies on the impact of different hyper-parameters}. for each hyper-parameter, the best value is highlighted in bold. Also, the default setting for reported results in the main context is marked with ``*".} 
    \label{tab:ablation_hyperparameter}
    \begin{tabular}{@{}c@{\hskip 3pt}|@{\hskip 3pt}c@{\hskip 3pt}|@{\hskip 3pt}c@{}}
 \small{Hyper-parameter} & \small{Image-level} & \small{Pixel-level}\\
\small{value} & \tiny{AUROC\,/\,AUPRC\,/\,f1$_{\text{mask}}$} & \tiny{AUROC\,/\,AUPRC\,/\,f1$_{\text{mask}}$\,/\,AUPRO} \\
 
\toprule

\multicolumn{3}{c}{\small{Patch Size set}}\\
\midrule
 \small{\phz$[1,2,4]$} & \imgperf{98.7}{99.5}{98.0} & \pixperf{98.3}{74.2}{69.2}{94.1} \\
  \small{*$[1,2,4,8]$}  & \imgperf{99.3}{\textbf{99.8}}{98.5} & \pixperf{98.4}{74.9}{69.7}{\textbf{94.9}} \\
  \small{\phz$[1,2,4,8,16]$} & \imgperf{\textbf{99.3}}{99.7}{\textbf{98.7}} & \pixperf{\textbf{98.7}}{\textbf{75.6}}{\textbf{70.2}}{94.3} \\
\midrule

\multicolumn{3}{c}{\small{Masking Ratio}}\\
\midrule
 \small{\phz$R_{\textit{mask}}:0.3$} & \imgperf{98.7}{99.5}{98.3} & \pixperf{97.6}{72.4}{68.0}{93.7} \\
  \small{\phz$R_{\textit{mask}}:0.5$} & \imgperf{99.1}{99.6}{98.5} & \pixperf{98.4}{74.7}{69.6}{94.7} \\
  \small{*$R_{\textit{mask}}:0.7$}  & \imgperf{\textbf{99.3}}{\textbf{99.8}}{\textbf{98.5}} & \pixperf{\textbf{98.4}}{\textbf{74.9}}{\textbf{69.7}}{\textbf{94.9}} \\

\midrule
\multicolumn{3}{c}{\small{Shuffled Patches Ratio}}\\
\midrule
 \small{\phz$R_{\textit{shuffle}}:0.0$} & \imgperf{99.1}{99.7}{98.5} & \pixperf{\textbf{98.6}}{\textbf{75.9}}{\textbf{70.5}}{94.8} \\
\small{*$R_{\textit{shuffle}}:0.3$}  & \imgperf{\textbf{99.3}}{\textbf{99.8}}{\textbf{98.5}} & \pixperf{98.4}{74.9}{69.7}{\textbf{94.9}} \\
\small{\phz$R_{\textit{shuffle}}:0.6$} & \imgperf{98.9}{99.6}{98.1} & \pixperf{98.6}{75.6}{70.3}{94.6} \\

\midrule
\multicolumn{3}{c}{\small{$\gamma_{p}$ and $\gamma_{l}$}}\\
\midrule
\small{\phz$\gamma_{p}:0.2$ - $\gamma_{l}:0.4$} & \imgperf{99.1}{99.7}{98.5} & \pixperf{98.4}{75.0}{69.8}{94.8} \\
\small{\phz$\gamma_{p}:0.4$ - $\gamma_{l}:0.2$} & \imgperf{99.1}{99.7}{98.3} & \pixperf{\textbf{98.4}}{\textbf{75.8}}{\textbf{70.4}}{\textbf{94.9}} \\
\small{*$\gamma_{p}:0.4$ - $\gamma_{l}:0.4$} & \imgperf{99.3}{99.8}{98.5} & \pixperf{98.4}{74.9}{69.7}{94.9} \\
\small{\phz$\gamma_{p}:0.4$ - $\gamma_{l}:0.6$} & \imgperf{99.3}{99.8}{98.5} & \pixperf{98.4}{74.2}{69.4}{94.4} \\
\small{\phz$\gamma_{p}:0.4$ - $\gamma_{l}: $ \, \xmark \, } & \imgperf{99.3}{99.8}{98.5} & \pixperf{98.4}{73.2}{69.0}{93.9} \\
\small{\phz$\gamma_{p}:0.6$ - $\gamma_{l}:0.4$} & \imgperf{99.3}{99.8}{98.5} & \pixperf{98.4}{74.7}{69.7}{94.4} \\
\small{\phz$\gamma_{p}: $ \, \xmark \, - $\gamma_{l}:0.4$} & \imgperf{99.3}{99.8}{98.5} & \pixperf{98.4}{74.3}{69.6}{93.7} \\
\small{\phz$\gamma_{p}: $ \, \xmark \, - $\gamma_{l}: $ \, \xmark \, } & \imgperf{\textbf{99.3}}{\textbf{99.8}}{\textbf{98.5}} & \pixperf{98.3}{72.8}{68.9}{93.5} \\
\bottomrule
\end{tabular}
\end{table}

\section{Additional Qualitative Results.}
\label{sec:more-qualitatives}
\begin{figure*}[!]
\centering
\includegraphics[width=0.9\linewidth]{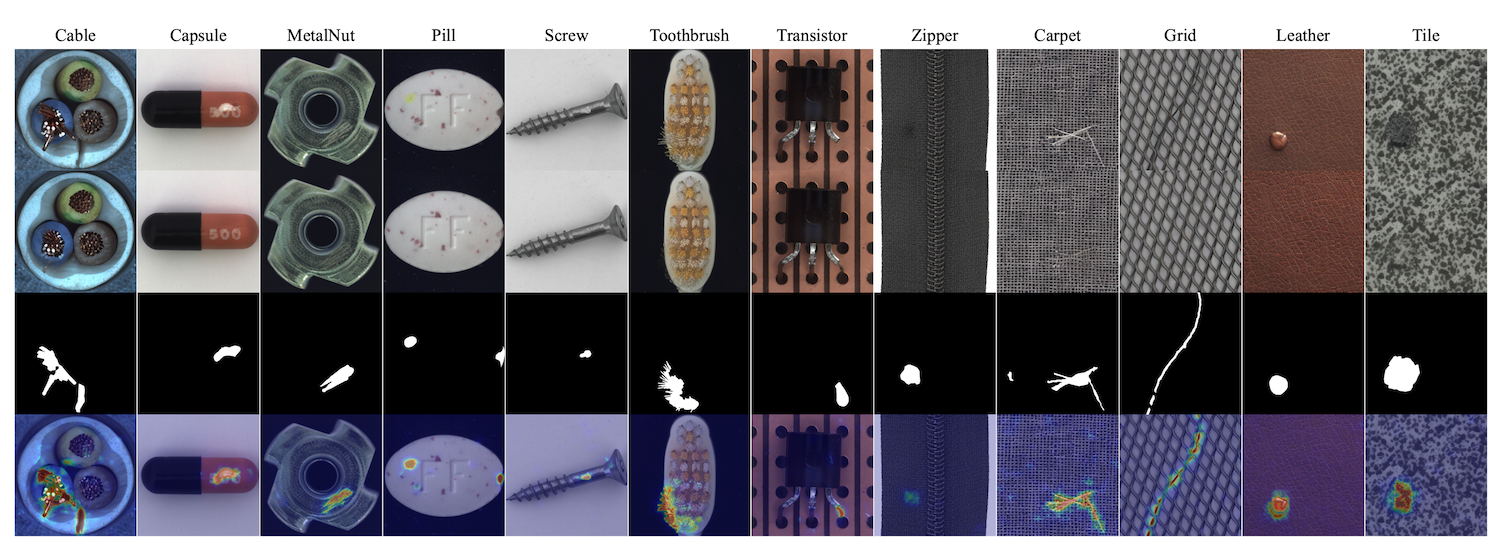}
     \caption{\textbf{Additional qualitative results on \MVTEC{} dataset.} From \textit{top} to \textit{bottom}: the original input image (with anomalies), \Ours{} reconstruction, the ground truth mask, and the predicted anomaly mask across different objects of \MVTEC{} dataset.}
\label{fig:qualitative_additional_mvtec}
\end{figure*}

\begin{figure*}[!]
\centering
\includegraphics[width=0.9\linewidth]{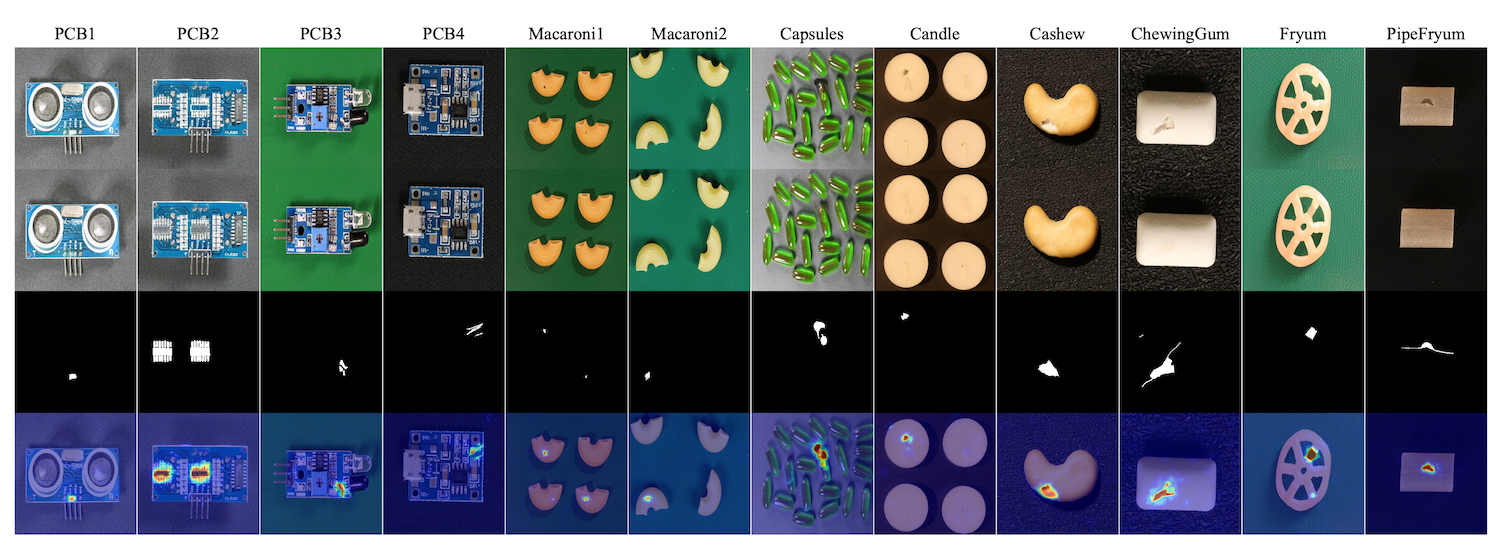}
     \caption{\textbf{Additional qualitative results on VisA dataset.} From \textit{top} to \textit{bottom}: the original input image (with anomalies), \Ours{} reconstruction, the ground truth mask, and the predicted anomaly mask across different objects of VisA dataset.}
\label{fig:qualitative_additional_visa}
\end{figure*}

\begin{figure*}[!]
\centering
\includegraphics[width=0.9\linewidth]{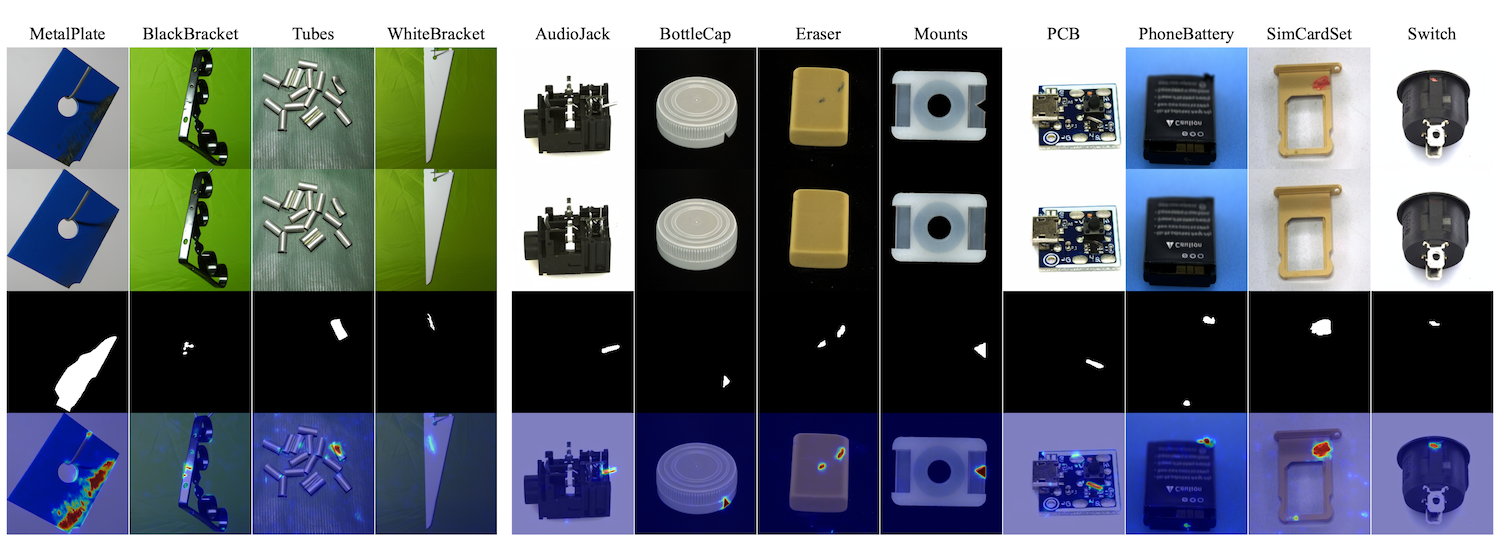}
     \caption{\textbf{Qualitative results on Additional datasets.} From \textit{top} to \textit{bottom}: the original input image (with anomalies), \Ours{} reconstruction, the ground truth mask, and the predicted anomaly mask for MPDD dataset (left side), and Real-IAD dataset(right side).}
\label{fig:qualitative_new_datasets}
\end{figure*}

We have visualized additional qualitative results for datasets, on 12 classes of \MVTEC{} and VisA dataset, respectively in \cref{fig:qualitative_additional_mvtec} and \cref{fig:qualitative_additional_visa} to further support the effectiveness and superiority of the proposed \Ours{} model. Also, we have depicted the results for MPDD and Real-IAD datasets respectively in \cref{fig:qualitative_new_datasets}.

\section{Limitations.}
\begin{figure*}[!]
\centering
\includegraphics[width=0.9\linewidth]{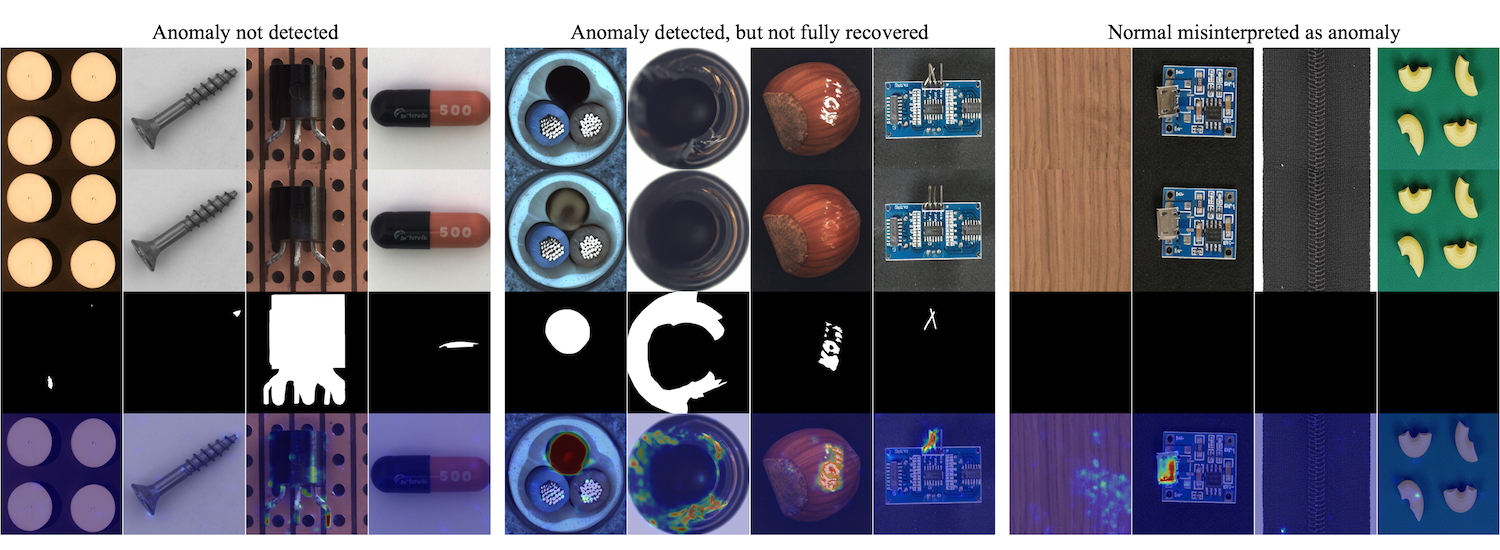}
     \caption{\textbf{Example of Failures on \MVTEC{} and Visa datasets.} Failures are categorized into three subsets, i.e. "Anomaly not detected", "Anomaly detected but not fully recovered", and "normal misinterpreted as an anomaly." from right to left respectively. For each image, from \textit{top} to \textit{bottom}: the original input image, \Ours{} reconstruction, the ground truth mask, and the predicted anomaly are depicted.}
\label{fig:failures}
\end{figure*}

In this section, we further explore the limitations of the proposed method, as well as analyzing the failures. For this purpose, we have visualized the failures of the methods in \cref{fig:failures}, and categorized them into three different subsets, i.e. \textit{``anomaly not detected"}, \textit{``anomaly detected but not fully recovered"}, and \textit{``normal detected as anomaly"},  First of all, we have considered the anomalies in the latent space as noise, while it might be too optimistic. This limitation becomes particularly pronounced when dealing with very large anomalies, which could result in not detecting the large displacements (as ``Transistor: in the third column), or not fully recovering the normal counterpart of the image (as ``bottle" in the sixth column). Exposing the model to more structured anomalies like synthetic anomalies could serve as a solution for this limitation. Also, as we have used VAE to map the images into the latent space, very small anomalies, like scratches that are barely visible, could be misinterpreted as the variation caused by variance of the VAE model, and therefore not detected. The first, second, and, fourth columns in \cref{fig:failures}, are failure examples due to this limitation. Training a better VAE model that is sensitive to these kinds of variations would probably improve the failures caused by this limitation. 
Furthermore, as \Ours{} model corrects the deviation progressively upon a point that the model considers them as in-distribution, in a few cases, there might still be the footprints of the anomaly, as is the case for the fifth, seventh, and eighth columns. This problem could be addressed by directly applying $\tilde{\zz}_{0}$ in each reverse time-step as proposed in Sec. 5.3.2 of the main paper, albeit at the cost of a slight decrease in image-level metrics.

\begin{table*}[!b]
    \centering
    \caption{\textbf{Image-level performance on \MVTEC.} Comparison to state-of-the-art methods on multi-class anomaly detection on the \MVTEC{} dataset. $\mathrm{AUROC\,/\,AUPRC\,/\,F1_{\max}}$ are reported.}
    \label{tab:img-mvtec}
    \resizebox{.95\textwidth}{!}{
    \begin{tabular}{lc|c|c|c|c|c|c|c|c}

\toprule
\rowcolor{white} & & UniAD \cite{you2022unified} & SimpleNet \cite{liu2023simplenet} & DeSTSeg \cite{zhang2023destseg} & DiAD \cite{he2024diffusion} & MambaAD \cite{he2024mambaad} & MoEAD \cite{mengmoead} & GLAD \cite{yao2024glad} & Deviation Correction \\
 & \multirow{-2}{*}{Category} & \footnotesize\textsl{NeurIPS'22} & \footnotesize\textsl{CVPR'23} & \footnotesize\textsl{CVPR'23} & \footnotesize\textsl{AAAI'24} & \footnotesize\textsl{NeurIPS'24} & \footnotesize\textsl{ECCV'24} & \footnotesize\textsl{ECCV'24}  & \footnotesize\textsl{Ours}\\
 \midrule
& Bottle &\imgperf{99.7}{\first{100.}}{\first{100.}} & \imgperf{\first{100.}}{\first{100.}}{\first{100.}} & \imgperf{98.7}{99.6}{96.8} & \imgperf{99.7}{96.5}{91.8} & \imgperf{\first{100.}}{\first{100.}}{\first{100.}} & \imgperf{\first{100.}}{\first{100.}}{\first{100.}} & \imgperf{\first{100.}}{\first{100.}}{\first{100.}} & \imgperf{\first{100.}}{\first{100.}}{\first{100.}} \\
& Cable &\imgperf{95.2}{95.9}{88.0} & \imgperf{97.5}{98.5}{94.7} & \imgperf{89.5}{94.6}{85.9} & \imgperf{94.8}{98.8}{95.2} & \imgperf{\second{98.8}}{\second{99.2}}{\second{95.7}} & \imgperf{98.7}{\second{99.2}}{\first{95.8}} & \imgperf{97.4}{98.6}{94.2} & \imgperf{\first{98.9}}{\first{99.3}}{95.6} \\
& Capsule &\imgperf{86.9}{97.8}{94.4} & \imgperf{90.7}{97.9}{93.5} & \imgperf{82.8}{95.9}{92.6} & \imgperf{89.0}{97.5}{95.5} & \imgperf{94.4}{98.7}{94.9} & \imgperf{93.7}{98.5}{96.4} & \imgperf{\first{96.6}}{\first{99.3}}{\second{96.8}} & \imgperf{\second{96.3}}{\second{99.2}}{\first{96.9}} \\
& Hazelnut &\imgperf{99.8}{\first{100.}}{99.3} & \imgperf{99.9}{99.9}{99.3} & \imgperf{98.8}{99.2}{98.6} & \imgperf{99.5}{99.7}{97.3} & \imgperf{\first{100.}}{\first{100.}}{\first{100.}} & \imgperf{\first{100.}}{\first{100.}}{\first{100.}} & \imgperf{97.1}{98.4}{94.5} & \imgperf{99.6}{99.8}{97.8} \\
& MetalNut &\imgperf{99.2}{99.9}{\first{99.5}} & \imgperf{96.9}{99.3}{96.1} & \imgperf{92.9}{98.4}{92.2} & \imgperf{99.1}{96.0}{91.6} & \imgperf{\second{99.9}}{\first{100.}}{\first{99.5}} & \imgperf{99.8}{\first{100.}}{98.9} & \imgperf{\first{100.}}{\first{100.}}{\first{99.5}} & \imgperf{99.7}{99.9}{98.9} \\
& Pill &\imgperf{93.7}{98.7}{95.7} & \imgperf{88.2}{97.7}{92.5} & \imgperf{77.1}{94.4}{91.7} & \imgperf{95.7}{98.5}{94.5} & \imgperf{\second{97.0}}{\second{99.5}}{\second{96.2}} & \imgperf{94.5}{98.9}{95.6} & \imgperf{95.5}{99.2}{95.0} & \imgperf{\first{99.6}}{\first{99.9}}{\first{99.6}} \\
& Screw  &\imgperf{87.5}{96.5}{89.0} & \imgperf{76.7}{90.6}{87.7} & \imgperf{69.9}{88.4}{85.4} & \imgperf{90.7}{\first{99.7}}{\first{97.9}} & \imgperf{94.7}{97.9}{94.0} & \imgperf{92.8}{97.4}{91.4} & \imgperf{\second{94.9}}{98.3}{93.7} & \imgperf{\first{99.1}}{\first{99.7}}{\first{97.9}} \\
& Toothbrush  &\imgperf{94.2}{97.4}{95.2} & \imgperf{89.7}{95.7}{92.3} & \imgperf{71.7}{89.3}{84.5} & \imgperf{\second{99.7}}{\second{99.9}}{\second{99.2}} & \imgperf{98.3}{99.3}{98.4} & \imgperf{95.0}{97.8}{96.8} & \imgperf{\second{99.7}}{\second{99.9}}{98.4} & \imgperf{\first{100.}}{\first{100.}}{\first{100.}} \\
& Transistor &\imgperf{99.8}{98.0}{93.8} & \imgperf{99.2}{98.7}{97.6} & \imgperf{78.2}{79.5}{68.8} & \imgperf{99.8}{99.6}{97.4} & \imgperf{\first{100.}}{\first{100.}}{\first{100.}} & \imgperf{99.8}{99.7}{97.5} & \imgperf{99.7}{99.6}{97.5} & \imgperf{\first{100.}}{\first{100.}}{\first{100.}} \\
\multirow{-10}{*}{\rotatebox[origin=c]{90}{Objects}} &  Zipper &\imgperf{95.8}{99.5}{97.1} & \imgperf{99.0}{99.7}{\second{98.3}} & \imgperf{88.4}{96.3}{93.1} & \imgperf{95.1}{99.1}{94.4} & \imgperf{\second{99.3}}{\second{99.8}}{97.5} & \imgperf{98.3}{99.5}{97.5} & \imgperf{97.9}{99.4}{96.3} & \imgperf{\first{99.8}}{\first{100.}}{\first{99.2}} \\
\midrule
& Carpet &\imgperf{\first{99.8}}{\first{99.9}}{\first{99.4}} & \imgperf{95.7}{98.7}{93.2} & \imgperf{95.9}{98.8}{94.9} & \imgperf{99.4}{\first{99.9}}{98.3} & \imgperf{\first{99.8}}{\first{99.9}}{\first{99.4}} & \imgperf{\first{99.8}}{\first{99.9}}{\first{99.4}} & \imgperf{96.8}{99.0}{95.6} & \imgperf{\first{99.8}}{\first{99.9}}{98.9} \\
& Grid  &\imgperf{98.2}{99.5}{97.3} & \imgperf{97.6}{99.2}{96.4} & \imgperf{97.9}{99.2}{96.6} & \imgperf{98.5}{99.8}{97.7} & \imgperf{\first{100.}}{\first{100.}}{\first{100.}} & \imgperf{99.1}{99.7}{98.2} & \imgperf{99.8}{99.9}{99.1} & \imgperf{\first{100.}}{\first{100.}}{\first{100.}} \\
& Leather &\imgperf{\first{100.}}{\first{100.}}{\first{100.}} & \imgperf{\first{100.}}{\first{100.}}{\first{100.}} & \imgperf{99.2}{99.8}{98.9} & \imgperf{99.8}{99.7}{97.6} & \imgperf{\first{100.}}{\first{100.}}{\first{100.}} & \imgperf{\first{100.}}{\first{100.}}{\first{100.}} & \imgperf{99.1}{99.7}{97.8} & \imgperf{99.7}{99.9}{98.9} \\
& Tile &\imgperf{99.3}{99.8}{98.2} & \imgperf{99.3}{99.8}{\second{98.8}} & \imgperf{97.0}{98.9}{95.3} & \imgperf{96.8}{\second{99.9}}{98.4} & \imgperf{98.2}{99.3}{95.4} & \imgperf{\second{99.4}}{99.8}{97.6} & \imgperf{\first{99.9}}{\first{100.}}{\first{99.4}} & \imgperf{99.1}{99.7}{97.6} \\
\multirow{-5}{*}{\rotatebox[origin=c]{90}{Textures}} &  Wood  &\imgperf{98.6}{99.6}{96.6} & \imgperf{98.4}{99.5}{96.7} & \imgperf{\first{99.9}}{\first{100.}}{\second{99.2}} & \imgperf{\second{99.7}}{\first{100.}}{\first{100.}} & \imgperf{98.8}{99.6}{96.6} & \imgperf{98.8}{99.6}{96.7} & \imgperf{94.3}{98.2}{93.6} & \imgperf{97.3}{99.1}{95.9} \\
\midrule
\multicolumn{2}{c|}{Average} &\imgperf{96.5}{98.8}{96.2} & \imgperf{95.3}{98.4}{95.8} & \imgperf{89.2}{95.5}{91.6} & \imgperf{97.2}{99.0}{96.5} & \imgperf{\second{98.6}}{\second{99.6}}{\second{97.8}} & \imgperf{98.0}{99.3}{97.5} & \imgperf{97.5}{99.1}{96.6} & \imgperf{\first{99.3}}{\first{99.8}}{\first{98.5}} \\

\bottomrule
\end{tabular}
}
\end{table*}

\begin{table*}[b]
    \centering
    \caption{\textbf{Pixel-level performance.} Comparison to state-of-the-art methods on multi-class anomaly detection on the \MVTEC{} dataset. The following metrics are reported: $\mathrm{AUROC\,/\,AUPRC\,/\,f1_{\max}\,/\,AUPRO}$. For each category, the best method (per metric) is highlighted in \first{blue}, whereas \second{red} is used to denote the second-best method.}
    \label{tab:px-mvtec}
    \setlength{\tabcolsep}{4pt}
    \resizebox{\linewidth}{!}{
    \begin{tabular}{lc|c|c|c|c|c|c|c|c}
\toprule
\rowcolor{white} & & UniAD \cite{you2022unified} & SimpleNet \cite{liu2023simplenet} & DeSTSeg \cite{zhang2023destseg} & DiAD \cite{he2024diffusion} & MambaAD \cite{he2024mambaad} & MoEAD \cite{mengmoead} & GLAD \cite{yao2024glad} & Deviation Correction \\
 & \multirow{-2}{*}{Category} & \footnotesize\textsl{NeurIPS'22} & \footnotesize\textsl{CVPR'23} & \footnotesize\textsl{CVPR'23} & \footnotesize\textsl{AAAI'24} & \footnotesize\textsl{NeurIPS'24} & \footnotesize\textsl{ECCV'24} & \footnotesize\textsl{ECCV'24} & \footnotesize\textsl{Ours}\\
 \midrule

& Bottle &\pixperf{98.1}{66.0}{69.2}{93.1} & \pixperf{97.2}{53.8}{62.4}{89.0} & \pixperf{93.3}{61.7}{56.0}{67.5} & \pixperf{98.4}{52.2}{54.8}{86.6} & \pixperf{\first{98.8}}{79.7}{\second{76.7}}{\second{95.2}} & \pixperf{98.0}{69.4}{67.0}{93.6} & \pixperf{98.3}{\second{80.3}}{74.7}{\first{96.0}} & \pixperf{\second{98.7}}{\first{86.5}}{\first{79.0}}{95.0} \\
& Cable &\pixperf{97.3}{39.9}{45.2}{86.1} & \pixperf{96.7}{42.4}{51.2}{85.4} & \pixperf{89.3}{37.5}{40.5}{49.4} & \pixperf{96.8}{50.1}{\second{57.8}}{80.5} & \pixperf{95.8}{42.2}{48.1}{\second{90.3}} & \pixperf{\second{97.7}}{\second{56.8}}{49.1}{89.6} & \pixperf{94.1}{52.9}{54.4}{89.4} & \pixperf{\first{98.4}}{\first{77.8}}{\first{71.1}}{\first{92.9}} \\
& Capsule &\pixperf{98.5}{42.7}{46.5}{92.1} & \pixperf{98.5}{35.4}{44.3}{84.5} & \pixperf{95.8}{47.9}{48.9}{62.1} & \pixperf{97.1}{42.0}{45.3}{87.2} & \pixperf{98.4}{43.9}{47.7}{92.6} & \pixperf{\second{98.6}}{48.4}{44.1}{90.2} & \pixperf{\first{99.1}}{\second{49.8}}{\second{52.2}}{\first{96.3}} & \pixperf{97.8}{\first{56.5}}{\first{55.9}}{\second{93.4}} \\
& Hazelnut &\pixperf{98.1}{55.2}{56.8}{94.1} & \pixperf{98.4}{44.6}{51.4}{87.4} & \pixperf{98.2}{65.8}{61.6}{84.5} & \pixperf{98.3}{\first{79.2}}{\first{80.4}}{91.5} & \pixperf{\first{99.0}}{63.6}{64.4}{\second{95.7}} & \pixperf{97.8}{54.4}{52.3}{92.3} & \pixperf{\first{99.0}}{71.2}{66.7}{91.9} & \pixperf{98.6}{\second{75.9}}{\second{68.3}}{\first{95.8}} \\
& MetalNut  &\pixperf{62.7}{14.6}{29.2}{81.8} & \pixperf{\second{98.0}}{\second{83.1}}{79.4}{85.2} & \pixperf{84.2}{42.0}{22.8}{53.0} & \pixperf{97.3}{30.0}{38.3}{90.6} & \pixperf{96.7}{74.5}{79.1}{93.7} & \pixperf{94.8}{68.0}{58.4}{88.5} & \pixperf{97.3}{81.2}{\second{82.3}}{\second{94.2}} & \pixperf{\first{98.3}}{\first{89.0}}{\first{84.9}}{\first{94.4}} \\
& Pill &\pixperf{95.0}{44.0}{53.9}{95.3} & \pixperf{96.5}{72.4}{67.7}{81.9} & \pixperf{96.2}{61.7}{41.8}{27.9} & \pixperf{95.7}{46.0}{51.4}{89.0} & \pixperf{97.4}{64.0}{66.5}{\second{95.7}} & \pixperf{95.8}{49.9}{40.8}{95.1} & \pixperf{\second{97.8}}{\second{73.9}}{\second{69.4}}{94.7} & \pixperf{\first{99.1}}{\first{81.1}}{\first{78.3}}{\first{96.9}} \\
& Screw &\pixperf{98.3}{28.7}{37.6}{95.2} & \pixperf{96.5}{15.9}{23.2}{84.0} & \pixperf{93.8}{19.9}{25.3}{47.3} & \pixperf{97.9}{\second{60.6}}{\second{59.6}}{95.0} & \pixperf{99.5}{49.8}{50.9}{\second{97.1}} & \pixperf{98.8}{37.1}{28.5}{95.1} & \pixperf{\second{99.6}}{47.9}{48.3}{96.6} & \pixperf{\first{99.8}}{\first{69.7}}{\first{65.5}}{\first{98.4}} \\
& Toothbrush &\pixperf{98.4}{34.9}{45.7}{87.9} & \pixperf{98.4}{46.9}{52.5}{87.4} & \pixperf{96.2}{52.9}{58.8}{30.9} & \pixperf{99.0}{\first{78.7}}{\first{72.8}}{95.0} & \pixperf{99.0}{48.5}{59.2}{91.7} & \pixperf{98.4}{49.6}{39.3}{87.7} & \pixperf{\second{99.2}}{47.1}{60.0}{\first{96.0}} & \pixperf{\first{99.4}}{\second{77.6}}{\second{72.3}}{\second{95.5}} \\
& Transistor &\pixperf{\first{97.9}}{59.5}{\second{64.6}}{\second{93.5}} & \pixperf{95.8}{58.2}{56.0}{83.2} & \pixperf{73.6}{38.4}{39.2}{43.9} & \pixperf{95.1}{15.6}{31.7}{90.0} & \pixperf{96.5}{\first{69.4}}{\first{67.1}}{87.0} & \pixperf{\second{97.6}}{\second{63.7}}{56.5}{\first{93.9}} & \pixperf{89.5}{55.9}{56.6}{86.1} & \pixperf{92.4}{57.8}{54.9}{87.4} \\
\multirow{-10}{*}{\rotatebox[origin=c]{90}{Objects}} & Zipper &\pixperf{96.8}{40.1}{49.9}{92.6} & \pixperf{97.9}{53.4}{54.6}{90.7} & \pixperf{97.3}{\second{64.7}}{59.2}{66.9} & \pixperf{96.2}{60.7}{60.0}{91.6} & \pixperf{\second{98.4}}{60.4}{\second{61.7}}{\second{94.3}} & \pixperf{97.7}{49.4}{39.2}{93.0} & \pixperf{92.9}{41.7}{47.1}{83.8} & \pixperf{\first{99.5}}{\first{84.1}}{\first{76.8}}{\first{97.9}} \\
\midrule
& Carpet &\pixperf{98.5}{49.9}{51.1}{94.4} & \pixperf{97.4}{38.7}{43.2}{90.6} & \pixperf{93.6}{59.9}{58.9}{89.3} & \pixperf{98.6}{42.2}{46.4}{90.6} & \pixperf{\second{99.2}}{60.0}{63.3}{\second{96.7}} & \pixperf{98.2}{50.1}{46.6}{94.0} & \pixperf{98.8}{\second{71.9}}{\second{68.3}}{95.0} & \pixperf{\first{99.3}}{\first{82.0}}{\first{74.6}}{\first{97.2}} \\
& Grid &\pixperf{63.1}{10.7}{11.9}{92.9} & \pixperf{96.8}{20.5}{27.6}{88.6} & \pixperf{97.0}{42.1}{46.9}{86.8} & \pixperf{96.6}{\first{66.0}}{\first{64.1}}{94.0} & \pixperf{99.2}{47.4}{47.7}{97.0} & \pixperf{97.4}{27.4}{22.3}{91.7} & \pixperf{\second{99.4}}{40.8}{45.3}{\second{97.6}} & \pixperf{\first{99.5}}{\second{63.9}}{\second{61.0}}{\first{97.9}} \\
& Leather &\pixperf{98.8}{32.9}{34.4}{96.8} & \pixperf{98.7}{28.5}{32.9}{92.7} & \pixperf{\second{99.5}}{\first{71.5}}{\first{66.5}}{91.1} & \pixperf{98.8}{56.1}{\second{62.3}}{91.3} & \pixperf{99.4}{50.3}{53.3}{\first{98.7}} & \pixperf{98.6}{31.7}{30.1}{96.7} & \pixperf{\first{99.7}}{\second{62.2}}{61.2}{\second{97.0}} & \pixperf{99.3}{58.8}{58.7}{95.4} \\
& Tile  &\pixperf{91.8}{42.1}{50.6}{78.4} & \pixperf{95.7}{60.5}{59.9}{90.6} & \pixperf{93.0}{71.0}{66.2}{87.1} & \pixperf{92.4}{65.7}{64.1}{90.7} & \pixperf{93.8}{45.1}{54.8}{80.0} & \pixperf{91.6}{50.4}{42.6}{78.8} & \pixperf{\second{97.9}}{\second{72.8}}{\second{75.1}}{\first{96.6}} & \pixperf{\first{98.2}}{\first{86.4}}{\first{76.5}}{\second{91.4}} \\
\multirow{-5}{*}{\rotatebox[origin=c]{90}{Textures}} & Wood &\pixperf{93.2}{37.2}{41.5}{86.7} & \pixperf{91.4}{39.7}{34.8}{76.3} & \pixperf{95.9}{\first{77.3}}{\first{71.3}}{83.4} & \pixperf{93.3}{43.3}{43.5}{\first{97.5}} & \pixperf{94.4}{46.2}{48.2}{91.2} & \pixperf{92.8}{39.9}{35.1}{85.1} & \pixperf{\second{96.8}}{68.6}{63.1}{86.7} & \pixperf{\first{97.6}}{\second{75.9}}{\second{68.4}}{\second{93.5}} \\
\midrule
\multicolumn{2}{c|}{Average} &\pixperf{96.8}{43.4}{49.5}{90.7} & \pixperf{96.9}{45.9}{49.7}{86.5} & \pixperf{93.1}{54.3}{50.9}{64.8} & \pixperf{96.8}{52.6}{55.5}{90.7} & \pixperf{\second{97.7}}{56.3}{59.2}{\second{93.1}} & \pixperf{96.9}{49.8}{43.5}{91.0} & \pixperf{97.4}{\second{60.8}}{\second{60.7}}{93.0} & \pixperf{\first{98.4}}{\first{74.9}}{\first{69.7}}{\first{94.9}} \\
 \bottomrule
\end{tabular}
}
\end{table*}

\begin{table*}[!]
\centering
\caption{\textbf{Image-level performance on VisA.} Comparison to state-of-the-art methods on multi-class anomaly detection on the VisA dataset. $\mathrm{AUROC\,/\,AUPRC\,/\,f1_{\max}}$ are reported.}
\label{tab:img-visa}
\setlength{\tabcolsep}{5pt}
\resizebox{.9\linewidth}{!}{
\begin{tabular}{lc|c|c|c|c|c|c|c|c}
\toprule
\rowcolor{white} & & UniAD \cite{you2022unified} & SimpleNet \cite{liu2023simplenet} & DeSTSeg \cite{zhang2023destseg} & DiAD \cite{he2024diffusion} & MambaAD \cite{he2024mambaad} & MoEAD \cite{mengmoead} & GLAD \cite{yao2024glad} & Deviation Correction \\
& \multirow{-2}{*}{Category} & \footnotesize\textsl{NeurIPS'22} & \footnotesize\textsl{CVPR'23} & \footnotesize\textsl{CVPR'23} & \footnotesize\textsl{AAAI'24} & \footnotesize\textsl{NeurIPS'24} & \footnotesize\textsl{ECCV'24} &\footnotesize\textsl{ECCV'24} & \footnotesize\textsl{Ours} \\
 \midrule

& PCB1 &\imgperf{92.8}{92.7}{87.8} & \imgperf{91.6}{91.9}{86.0} & \imgperf{87.6}{83.1}{83.7} & \imgperf{88.1}{88.7}{80.7} & \imgperf{95.4}{93.0}{91.6} & \imgperf{\first{97.7}}{\first{97.3}}{\first{95.2}} & \imgperf{78.1}{79.8}{74.0} & \imgperf{\second{96.7}}{\second{95.9}}{\second{94.6}} \\
& PCB2 &\imgperf{87.8}{87.7}{83.1} & \imgperf{92.4}{93.3}{84.5} & \imgperf{86.5}{85.8}{82.6} & \imgperf{91.4}{91.4}{84.7} & \imgperf{94.2}{93.7}{89.3} & \imgperf{\second{95.1}}{\second{95.6}}{\second{90.1}} & \imgperf{88.0}{86.8}{82.8} & \imgperf{\first{97.2}}{\first{96.5}}{\first{93.3}} \\
& PCB3 &\imgperf{78.6}{78.6}{76.1} & \imgperf{89.1}{91.1}{82.6} & \imgperf{93.7}{95.1}{87.0} & \imgperf{86.2}{87.6}{77.6} & \imgperf{93.7}{94.1}{86.7} & \imgperf{92.2}{92.6}{85.2} & \imgperf{\second{95.9}}{\second{96.1}}{\second{87.7}} & \imgperf{\first{97.7}}{\first{97.9}}{\first{94.1}} \\
\multirow{-4}{*}{\rotatebox[origin=c]{90}{Complex}} & PCB4 &\imgperf{98.8}{98.8}{94.3} & \imgperf{97.0}{97.0}{93.5} & \imgperf{97.8}{97.8}{92.7} & \imgperf{99.6}{99.5}{97.0} & \imgperf{\first{99.9}}{\first{99.9}}{\first{98.5}} & \imgperf{\second{99.7}}{\second{99.7}}{97.0} & \imgperf{99.3}{99.1}{\second{97.5}} & \imgperf{98.5}{97.3}{94.6} \\
\midrule
& Macaroni1 &\imgperf{79.9}{79.8}{72.7} & \imgperf{85.9}{82.5}{73.1} & \imgperf{76.6}{69.0}{71.0} & \imgperf{85.7}{85.2}{78.8} & \imgperf{91.6}{89.8}{81.6} & \imgperf{\second{93.0}}{\second{93.2}}{85.8} & \imgperf{91.5}{91.7}{\second{86.2}} & \imgperf{\first{95.0}}{\first{95.4}}{\first{88.4}} \\
& Macaroni2 &\imgperf{71.6}{71.6}{69.9} & \imgperf{68.3}{54.3}{59.7} & \imgperf{68.9}{62.1}{67.7} & \imgperf{62.5}{57.4}{69.6} & \imgperf{81.6}{78.0}{73.8} & \imgperf{\second{86.3}}{\second{88.7}}{\first{80.4}} & \imgperf{73.8}{71.2}{71.8} & \imgperf{\first{88.4}}{\first{90.2}}{\first{80.4}} \\
& Capsules  &\imgperf{55.6}{55.6}{76.9} & \imgperf{74.1}{82.8}{74.6} & \imgperf{87.1}{93.0}{84.2} & \imgperf{58.2}{69.0}{78.5} & \imgperf{91.8}{95.0}{\second{88.8}} & \imgperf{77.6}{87.8}{79.7} & \imgperf{\second{92.4}}{\second{95.9}}{88.0} & \imgperf{\first{95.8}}{\first{97.7}}{\first{91.2}} \\
\multirow{-4}{*}{\rotatebox[origin=c]{90}{Multiple}} & Candle &\imgperf{94.1}{94.0}{86.1} & \imgperf{84.1}{73.3}{76.6} & \imgperf{94.9}{94.8}{89.2} & \imgperf{92.8}{92.0}{87.6} & \imgperf{\second{96.8}}{\second{96.9}}{\second{90.1}} & \imgperf{\first{97.2}}{\first{97.3}}{\first{92.8}} & \imgperf{88.1}{88.8}{81.8} & \imgperf{95.6}{95.5}{88.1} \\
\midrule
& Cashew  &\imgperf{92.8}{92.8}{91.4} & \imgperf{88.0}{91.3}{84.7} & \imgperf{92.0}{96.1}{88.1} & \imgperf{91.5}{95.7}{89.7} & \imgperf{94.5}{97.3}{91.1} & \imgperf{90.7}{95.3}{89.2} & \imgperf{\second{96.6}}{\first{98.5}}{\second{94.6}} & \imgperf{\first{97.8}}{\first{98.5}}{\first{95.6}} \\
& ChewingGum &\imgperf{96.3}{96.2}{95.2} & \imgperf{96.4}{98.2}{93.8} & \imgperf{95.8}{98.3}{94.7} & \imgperf{\second{99.1}}{99.5}{95.9} & \imgperf{97.7}{98.9}{94.2} & \imgperf{98.9}{\second{99.6}}{\first{98.5}} & \imgperf{\first{99.3}}{\first{99.7}}{\second{97.0}} & \imgperf{97.6}{98.7}{93.3} \\
& Fryum &\imgperf{83.0}{83.0}{85.0} & \imgperf{88.4}{93.0}{83.3} & \imgperf{92.1}{96.1}{89.5} & \imgperf{89.8}{95.0}{87.2} & \imgperf{95.2}{97.7}{90.5} & \imgperf{90.8}{95.8}{88.4} & \imgperf{\second{98.8}}{\second{99.4}}{\second{96.6}} & \imgperf{\first{99.3}}{\first{99.6}}{\first{98.7}} \\
\multirow{-4}{*}{\rotatebox[origin=c]{90}{Single}} & PipeFryum  &\imgperf{94.7}{94.7}{93.9} & \imgperf{90.8}{95.5}{88.6} & \imgperf{94.1}{97.1}{91.9} & \imgperf{96.2}{98.1}{93.7} & \imgperf{\second{98.7}}{\second{99.3}}{\second{97.0}} & \imgperf{96.7}{98.4}{95.0} & \imgperf{\first{99.7}}{\first{99.9}}{\first{98.0}} & \imgperf{97.7}{98.6}{94.5} \\
\midrule
& Average &\imgperf{85.5}{85.5}{84.4} & \imgperf{87.2}{87.0}{81.8} & \imgperf{88.9}{89.0}{85.2} & \imgperf{86.8}{88.3}{85.1} & \imgperf{\second{94.3}}{94.5}{89.4} & \imgperf{93.0}{\second{95.1}}{\second{89.8}} & \imgperf{91.8}{92.2}{88.0} & \imgperf{\first{96.4}}{\first{96.8}}{\first{92.2}} \\
\bottomrule
\end{tabular}
}
\end{table*}

\begin{table*}[!]
\centering
\caption{\textbf{Pixel-level performance.} Comparison to state-of-the-art methods on multi-class anomaly detection on the VisA dataset. The following metrics are reported: $\mathrm{AUROC\,/\,AUPRC\,/\,f1_{\max}\,/\,AUPRO}$. For each category, the best method (per metric) is highlighted in \first{blue}, whereas \second{red} is used to denote the second-best method.}
\label{tab:px-visa}
\setlength{\tabcolsep}{5pt}
\resizebox{\linewidth}{!}{
\begin{tabular}{lc|c|c|c|c|c|c|c|c}
\toprule
\rowcolor{white} & & UniAD \cite{you2022unified} & SimpleNet \cite{liu2023simplenet} & DeSTSeg \cite{zhang2023destseg} & DiAD \cite{he2024diffusion} & MambaAD \cite{he2024mambaad} & MoEAD \cite{mengmoead} & GLAD \cite{yao2024glad} & Deviation Correction \\
 & \multirow{-2}{*}{Category} & \footnotesize\textsl{NeurIPS'22} & \footnotesize\textsl{CVPR'23} & \footnotesize\textsl{CVPR'23} & \footnotesize\textsl{AAAI'24} & \footnotesize\textsl{NeurIPS'24} & \footnotesize\textsl{ECCV'24} &\footnotesize\textsl{ECCV'24} & \footnotesize\textsl{Ours} \\
 \midrule

& PCB1 &\pixperf{93.3}{3.9}{8.3}{64.1} & \pixperf{99.2}{\first{86.1}}{\first{78.8}}{83.6} & \pixperf{95.8}{46.4}{49.0}{83.2} & \pixperf{98.7}{49.6}{52.8}{80.2} & \pixperf{\first{99.8}}{\second{77.1}}{\second{72.4}}{\second{92.8}} & \pixperf{\second{99.6}}{64.1}{68.2}{92.0} & \pixperf{97.5}{38.1}{45.9}{91.9} & \pixperf{99.5}{66.0}{69.8}{\first{94.0}} \\
& PCB2 &\pixperf{93.9}{4.2}{9.2}{66.9} & \pixperf{96.6}{8.9}{18.6}{85.7} & \pixperf{97.3}{14.6}{\second{28.2}}{79.9} & \pixperf{95.2}{7.5}{16.7}{67.0} & \pixperf{\second{98.9}}{13.3}{23.4}{89.6} & \pixperf{98.4}{\second{19.0}}{11.7}{86.0} & \pixperf{97.5}{5.4}{12.5}{\first{90.8}} & \pixperf{\first{99.1}}{\first{56.2}}{\first{55.3}}{\first{90.8}} \\
& PCB3  &\pixperf{97.3}{13.8}{21.9}{70.6} & \pixperf{97.2}{\second{31.0}}{\second{36.1}}{85.1} & \pixperf{97.7}{28.1}{33.4}{62.4} & \pixperf{96.7}{8.0}{18.8}{68.9} & \pixperf{\first{99.1}}{18.3}{27.4}{89.1} & \pixperf{\second{98.9}}{26.0}{25.0}{84.3} & \pixperf{97.0}{24.9}{27.6}{\first{95.3}} & \pixperf{98.7}{\first{49.1}}{\first{52.0}}{\second{90.1}} \\
\multirow{-4}{*}{\rotatebox[origin=c]{90}{Complex}} & PCB4 &\pixperf{94.9}{14.7}{22.9}{72.3} & \pixperf{93.9}{23.9}{32.9}{61.1} & \pixperf{95.8}{\first{53.0}}{\first{53.2}}{76.9} & \pixperf{97.0}{17.6}{27.2}{85.0} & \pixperf{\second{98.6}}{47.0}{46.9}{\second{87.6}} & \pixperf{97.8}{34.9}{29.4}{85.0} & \pixperf{\first{99.4}}{\second{52.2}}{\first{53.2}}{\first{94.6}} & \pixperf{96.3}{46.5}{44.2}{84.0} \\
\midrule
& Macaroni1 &\pixperf{97.4}{3.7}{9.7}{84.0} & \pixperf{98.9}{3.5}{8.4}{92.0} & \pixperf{99.1}{5.8}{13.4}{62.4} & \pixperf{94.1}{10.2}{16.7}{68.5} & \pixperf{99.5}{17.5}{27.6}{95.2} & \pixperf{99.5}{\second{21.5}}{11.9}{\second{96.5}} & \pixperf{\first{99.9}}{18.4}{\second{32.6}}{\first{99.2}} & \pixperf{\second{99.6}}{\first{42.0}}{\first{36.8}}{96.3} \\
& Macaroni2 &\pixperf{95.2}{0.9}{4.3}{76.6} & \pixperf{93.2}{0.6}{3.9}{77.8} & \pixperf{98.5}{6.3}{14.4}{70.0} & \pixperf{93.6}{0.9}{2.8}{73.1} & \pixperf{\second{99.5}}{9.2}{\second{16.1}}{\second{96.2}} & \pixperf{98.5}{\second{14.6}}{6.6}{91.4} & \pixperf{\first{99.6}}{5.7}{12.2}{\first{98.0}} & \pixperf{98.6}{\first{28.1}}{\first{24.7}}{\second{96.2}} \\
& Capsules &\pixperf{88.7}{3.0}{7.4}{43.7} & \pixperf{97.1}{52.9}{53.3}{73.7} & \pixperf{96.9}{33.2}{9.1}{76.7} & \pixperf{97.3}{10.0}{21.0}{77.9} & \pixperf{99.1}{\second{61.3}}{\second{59.8}}{91.8} & \pixperf{98.9}{58.4}{59.4}{80.6} & \pixperf{\second{99.3}}{48.4}{52.0}{\second{92.1}} & \pixperf{\first{99.8}}{\first{70.9}}{\first{71.0}}{\first{94.4}} \\
\multirow{-4}{*}{\rotatebox[origin=c]{90}{Multiple}} & Candle &\pixperf{98.5}{17.6}{27.9}{91.6} & \pixperf{97.6}{8.4}{16.5}{87.6} & \pixperf{98.7}{\first{39.9}}{\first{45.8}}{69.0} & \pixperf{97.3}{12.8}{22.8}{89.4} & \pixperf{99.0}{23.2}{32.4}{\first{95.5}} & \pixperf{\first{99.3}}{34.8}{25.7}{\second{94.6}} & \pixperf{98.9}{26.5}{34.2}{94.0} & \pixperf{\second{99.1}}{\second{37.0}}{\second{36.3}}{\second{94.6}} \\
\midrule
& Cashew &\pixperf{98.6}{51.7}{58.3}{87.9} & \pixperf{\second{98.9}}{\first{68.9}}{\first{66.0}}{84.1} & \pixperf{87.9}{47.6}{52.1}{66.3} & \pixperf{90.9}{53.1}{\second{60.9}}{61.8} & \pixperf{94.3}{46.8}{51.4}{87.8} & \pixperf{98.2}{50.3}{45.9}{\second{90.2}} & \pixperf{84.9}{24.1}{34.3}{60.3} & \pixperf{\first{99.0}}{\second{54.6}}{57.0}{\first{94.2}} \\
& ChewingGum  &\pixperf{98.8}{54.9}{56.1}{81.3} & \pixperf{97.9}{26.8}{29.8}{78.3} & \pixperf{98.8}{\first{86.9}}{\first{81.0}}{68.3} & \pixperf{94.7}{11.9}{25.8}{59.5} & \pixperf{98.1}{57.5}{59.9}{79.7} & \pixperf{99.3}{59.6}{59.3}{\second{84.1}} & \pixperf{\first{99.7}}{\second{78.5}}{73.1}{\first{93.3}} & \pixperf{\second{99.4}}{73.3}{\second{79.9}}{81.6} \\
& Fryum  &\pixperf{95.9}{34.0}{40.6}{76.2} & \pixperf{93.0}{39.1}{45.4}{85.1} & \pixperf{88.1}{35.2}{38.5}{47.7} & \pixperf{\first{97.6}}{\first{58.6}}{\first{60.1}}{81.3} & \pixperf{96.9}{47.8}{\second{51.9}}{91.6} & \pixperf{\second{97.4}}{\second{53.0}}{44.9}{84.1} & \pixperf{97.2}{39.8}{47.1}{\first{96.6}} & \pixperf{93.9}{45.9}{42.0}{\second{92.5}} \\
\multirow{-4}{*}{\rotatebox[origin=c]{90}{Single}} & PipeFryum  &\pixperf{98.9}{50.2}{57.7}{91.5} & \pixperf{98.5}{65.6}{63.4}{83.0} & \pixperf{98.9}{\first{78.8}}{\first{72.7}}{45.9} & \pixperf{\first{99.4}}{\second{72.7}}{\second{69.9}}{89.9} & \pixperf{99.1}{53.5}{58.5}{95.1} & \pixperf{99.0}{55.3}{51.3}{94.7} & \pixperf{99.1}{53.8}{59.1}{\first{98.4}} & \pixperf{\first{99.4}}{46.0}{45.0}{\second{96.3}} \\
\midrule
& Average &\pixperf{95.9}{21.0}{27.0}{75.6} & \pixperf{96.8}{34.7}{37.8}{81.4} & \pixperf{96.1}{\second{39.6}}{43.4}{67.4} & \pixperf{96.0}{26.1}{33.0}{75.2} & \pixperf{\second{98.5}}{39.4}{\second{44.0}}{91.0} & \pixperf{\first{98.7}}{36.6}{41.0}{88.6} & \pixperf{97.5}{34.6}{40.3}{\second{92.0}} & \pixperf{\second{98.5}}{\first{51.3}}{\first{51.2}}{\first{92.1}} \\

\bottomrule
\end{tabular}
}
\end{table*}

\begin{table*}[!]
    \centering
    \footnotesize
    \caption{\textbf{Quantitatve evaluation  on additional datasets.}. Image and Pixel-level results  
    on \textbf{MPDD} and \textbf{Real-IAD} datasets in \textit{multi-class UAD}. The best method (per metric) is highlighted in \first{blue}, whereas \second{red} is used to denote the second-best approach.}
    \label{tab:additional-datasets}
    \begin{tabular}{l|l | ccc | cccc}
\multicolumn{1}{c|}{\multirow{2}{*}{Dataset}} & \multicolumn{1}{c|}{\multirow{2}{*}{Method}} & \multicolumn{3}{c|}{Image-level} & \multicolumn{4}{c}{Pixel-level}\\
 & & \footnotesize{AUROC} & \footnotesize{AUPRC} & \footnotesize{f1$_{\text{max}}$} &  \footnotesize{AUROC} & \footnotesize{AUPRC} & \footnotesize{f1$_{\text{max}}$} & \footnotesize{AUPRO} \\
 \toprule

\multirow{9}{*}{\textbf{MPDD} \cite{jezek2021deep}} & RD4AD \cite{deng2022anomaly} \tiny\textsl{CVPR'22} & 90.3 & 92.8 & 90.5 & \first{98.3} & 39.6 & 40.6 & 95.2 \\ 
& UniAD \cite{you2022unified} \tiny\textsl{NeurIPS'22} & 80.1 & 83.2 & 85.1 & 95.4 & 19.0 & 25.6 & 83.8 \\ 
& SimpleNet \cite{liu2023simplenet} \tiny\textsl{CVPR'23} & 90.6 & 94.1 & 89.7 & 97.1 & 33.6 & 35.7 & 90.0 \\ 
& DeSTSeg \cite{zhang2023destseg} \tiny\textsl{CVPR'23} & 92.6 & 91.8 & 92.8 & 90.8 & 30.6 & 32.9 & 78.3 \\ 
& DiAD \cite{he2024diffusion} \tiny\textsl{AAAI'24} & 85.8 & 89.2 & 86.5 & 91.4 & 15.3 & 19.2 & 66.1 \\ 
& GLAD \cite{yao2024glad} \tiny\textsl{ECCV'24} & \second{97.5} & \second{97.1} & \first{96.8} & \second{98.0} & \second{40.9} & \second{41.5} & \first{93.0} \\ 
& MambaAD \cite{he2024mambaad} \tiny\textsl{NeurIPS'24} & 89.2 & 93.1 & 90.3 & 97.7 & 33.5 & 38.6 & \second{92.8} \\ 
\cmidrule{2-9}
& \textbf{\Ours{} (\textit{Ours})} &  \first{97.7} & \first{97.3} & \second{95.3} & 95.1 & \first{45.3} & \first{46.6} & 79.5\\ 
\midrule
\multirow{9}{*}{\textbf{Real-IAD} \cite{wang2024real}} & RD4AD \cite{deng2022anomaly} \tiny\textsl{CVPR'22} & 82.4 & 79.0 & 73.9 & 97.3 & 25.0 & 32.7 & \second{89.6} \\ 
& UniAD \cite{you2022unified} \tiny\textsl{NeurIPS'22} & 83.0 & 80.9 & 74.3 & 97.3 & 21.1 & 29.2 & 86.7  \\ 
& SimpleNet \cite{liu2023simplenet} \tiny\textsl{CVPR'23} & 57.2 & 53.4 & 61.5 & 75.7 & \phz2.8 & \phz6.5 & 39.0 \\ 
& DeSTSeg \cite{zhang2023destseg} \tiny\textsl{CVPR'23} & 82.3 & 79.2 & 73.2 & 94.6 & \second{37.9} & \second{41.7} & 40.6 \\ 
& DiAD \cite{he2024diffusion} \tiny\textsl{AAAI'24} & 75.6 & 66.4 & 69.9 & 88.0 & \phz2.9 & \phz7.1 & 58.1 \\ 
& MambaAD \cite{he2024mambaad} \tiny\textsl{NeurIPS'24} & \second{86.3} & \second{84.6} & \second{77.0} & \first{98.5} & 33.0 & 38.7 & \first{90.5} \\ 
\cmidrule{2-9}
& \textbf{\Ours{} (\textit{Ours})} & \first{87.0} & \first{86.1} & \first{79.2} & \second{97.4} & \first{46.4} & \first{48.6} & 88.8 \\ 

\bottomrule
\end{tabular}
\end{table*}

\end{document}